\documentclass[a4paper,fleqn]{cas-dc}

\usepackage[numbers]{natbib}
\usepackage{cancel}
\usepackage{ulem}
\usepackage{float}      
\usepackage{stfloats}   
\usepackage{placeins}   

\begin{document}
\fnbelowfloat
\let\WriteBookmarks\relax
\renewcommand{\floatpagefraction}{0.8}
\renewcommand{\textfraction}{0.05}
\renewcommand{\topfraction}{0.9}
\renewcommand{\bottomfraction}{0.9}
\setcounter{totalnumber}{10}
\setcounter{topnumber}{8}
\setcounter{bottomnumber}{8}

\shorttitle{Spectral filtering convolution neural network and high-pass kernel generation transformer}

\shortauthors{Sriprabha Ramanarayanan et~al.}

\title [mode = title]{SHFormer: Dynamic Spectral Filtering Convolutional Neural Network and High-pass Kernel Generation Transformer for Adaptive MRI Reconstruction}                      

\author[iitm,htic]{Sriprabha Ramanarayanan}
\cormark[1]
\author[iitm]{Rahul G. S.}
\author[iitm,htic]{Mohammad Al Fahim}
\author[htic]{Keerthi Ram}
\author[ge]{Ramesh Venkatesan}
\author[iitm,htic]{Mohanasankar Sivaprakasam}

\cortext[cor1]{Corresponding author}

\affiliation[iitm]{organization={Department of Electrical Engineering, Indian Institute of Technology Madras (IITM)},
            country={India}}

\affiliation[htic]{organization={Healthcare Technology Innovation Centre},
            addressline={IITM}, 
            country={India}}
\affiliation[ge]{organization={GE John F Welch Technology Center},
            addressline={GE Healthcare}, 
            country={India}}

\begin{abstract}
\LaTeX\ 
Attention Mechanism (AM) selectively focuses on essential information for imaging tasks and captures relationships between regions from distant pixel neighborhoods to compute feature representations. Accelerated magnetic resonance image (MRI) reconstruction can benefit from AM, as the imaging process involves acquiring Fourier domain measurements that influence the image representation in a non-local manner. However, AM-based models are more adept at capturing low-frequency information and have limited capacity in constructing high-frequency representations, restricting the models to smooth reconstruction. Secondly, AM-based models need mode-specific retraining for multimodal MRI data as their knowledge is restricted to local contextual variations within modes that might be inadequate to capture the diverse transferable features across heterogeneous data domains. To address these challenges, we propose a neuromodulation-based discriminative multi-spectral AM for scalable MRI reconstruction, that can (i) propagate the context-aware high-frequency details for high-quality image reconstruction, and (ii) capture features reusable to deviated unseen domains in multimodal MRI, to offer high practical value for the healthcare industry and researchers. The proposed network consists of a spectral filtering convolutional neural network to capture mode-specific transferable features to generalize to deviated MRI data domains and a dynamic high-pass kernel generation transformer that focuses on high-frequency details for improved reconstruction. We have evaluated our model on various aspects, such as comparative studies in supervised and self-supervised learning, diffusion model-based training, closed-set and open-set generalization under heterogeneous MRI data, and interpretation-based analysis. Our results show that the proposed method offers scalable and high-quality reconstruction with best improvement margins of $\sim$1 dB in PSNR and $\sim$0.01 in SSIM under unseen scenarios. Our code is available at \url{https://github.com/sriprabhar/SHFormer}.
\end{abstract}


\begin{highlights}
\item A neuromodulation-based discriminative multi-spectral attention mechanism consisting of a spectral filtering CNN and high-pass kernel generation transformer for scalable MRI reconstruction 
\item Instance-specific dynamic weight prediction to learn high-frequency details and capture features reusable to deviated unseen domains in multimodal MRI
\item Applicable to varying learning methods - supervised, self-supervised, and diffusion models
\item Extensive experimentation against other MRI reconstruction methods, adaptive models, and meta-learning methods with perspectives of closed and open-set generalization for cardiac, knee, and multi-contrast brain MRI
\end{highlights}


\maketitle

\section{Introduction}

\begin{figure}[t!]
    \centering
    \includegraphics[width=0.9\linewidth]{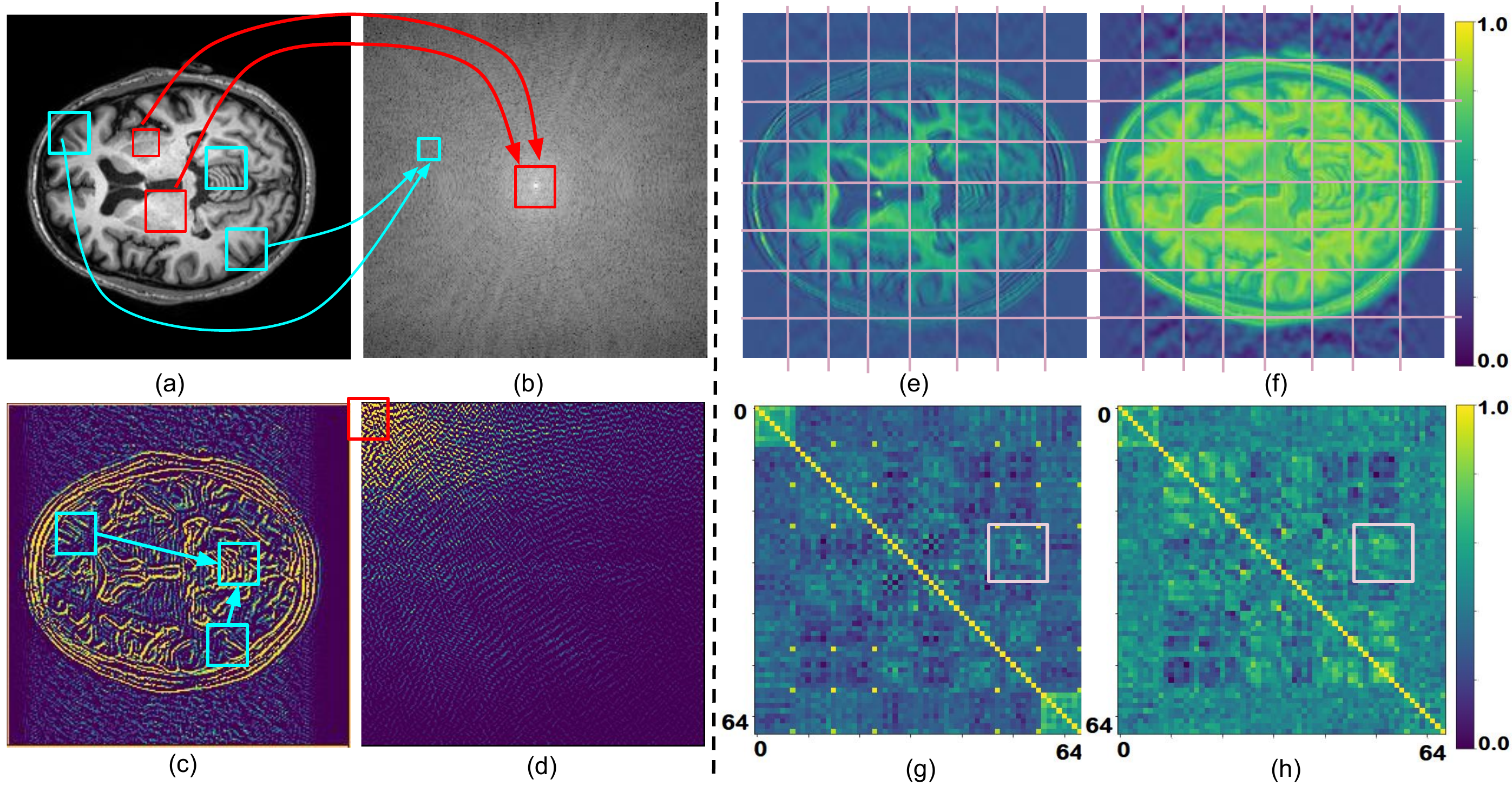}
    \caption{AM in MRI reconstruction. (a) Input image (b) k-space of (a). The lower k-space frequencies correspond to smooth variations 
    (red box) in the image while higher k-space frequencies correspond to fine details (cyan arrows).  Distant patches with  similar patterns share frequency components (cyan boxes in (a) and (b). (c) High-pass filtering AM features. (d) DCT-based spectral AM features. The AM computes the hidden representation patch-wise aggregating the low and high-frequency information from all other patches, contributing to the k-space components. A sublayer feature pair (e)  before AM, (f) after AM. A sample 64 $\times$ 64 normalized cross correlation (NCC) matrix (g)  before AM (h) after AM between every pair of  patches in the hidden features. The AM (h) exhibits higher distant patch similarity measures (pink box) compared to the case without AM (g).
    }
    \label{fig:AMforMRI}
\end{figure}

Attention mechanism (AM) is an important and challenging problem in feature modeling in deep learning that aims to focus on important information in features of a deep neural network \cite{attnallyouneed}. AMs have recently attracted considerable focus in deep neural networks (like convolutional neural networks or CNNs and transformers) for various tasks in medical imaging, computer vision, and natural language processing \cite{am_review}. CNNs adopt channel attention as an AM that directly learns to attach importance weights with different channels \cite{rcan}. The simplicity and efficacy of channel attention make it a popular and powerful tool for various deep learning-based image restoration tasks. The self-attention mechanism in vision transformers is another architectural breakthrough that enlarges the receptive field \cite{swin} and relates different patches within an image to capture relationships between regions from non-local or distant neighborhoods to compute feature representations \cite{swin,restomer}. Accelerated image reconstruction in the context of magnetic resonance imaging (MRI) can specifically benefit from the merits mentioned above of AM, as the imaging process includes acquiring measurements along k-space (Fourier domain signal)  trajectories, influencing the image domain representation in a non-local manner \cite{sdlformer} (Figure  \ref{fig:AMforMRI}).
Despite advancements, challenges are associated with these methods analyzed from the perspectives of training the model and the data domains.  

(i) \textbf{Dominance of low-frequencies components}: Previous deep neural network approaches are biased towards low-frequency signals \cite{HFComponentHelpsExplain, dff, mmca}. This observation might be due to the loss function used for training and the type of AM employed in the architecture. For instance, in the optimization process, utilizing conventional pixel-wise differences in the loss function might influence the reconstruction towards an average of the possible reconstructions that are equally distant in terms of $L_2$ loss in the higher dimensional manifold of images \cite{mmca}. The residual channel AM \cite{rcan} for CNN aggregates features using global average pooling capturing the lowest frequency component, discarding all higher frequencies, which are crucial for reconstruction. 
Moreover, the self-attention mechanism-based transformers are more adept at capturing low-frequency information and have limited capacity in constructing high-frequency representations \cite{freq_modulation_transformer}. These limitations restrict the models to smooth reconstructed image predictions, where there are scopes to improve the recovery of high-frequency details. 

\begin{figure*}[t!]
    \centering
    \includegraphics[width=0.9\linewidth]{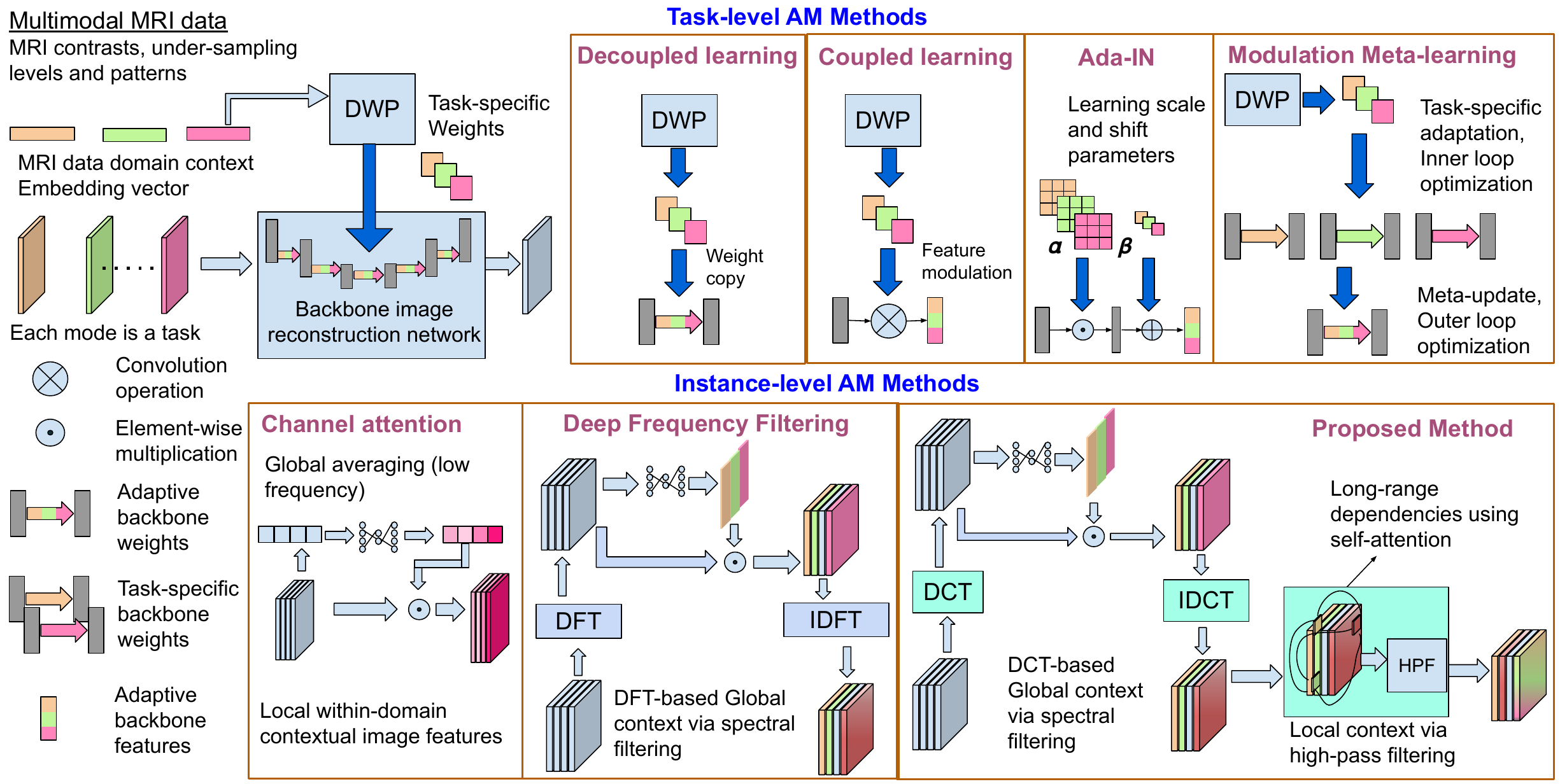}
    \caption{Concept diagram comparing previous neuromodulation-based AMs with the proposed method for heterogeneous MRI data. Methods like decoupled learning \cite{mac}, coupled learning \cite{mcihypernet}, and kernel modulation meta-learning \cite{kmmaml} use a dynamic weight prediction (DWP) network with the backbone image reconstruction network. These methods extract task-specific features, wherein each mode of the multimodal MRI data is posed as a task. Methods like adaptive instance normalization (Ada-IN) \cite{univusmri}, channel attention \cite{miccan}, and deep frequency filtering \cite{dff} operate at the instance level, focusing on either local contextual information within images or global task-level or mode-specific information. All these methods incline toward low-frequency components of the image data. The proposed approach captures both global task-level and instance-level features with a focus on high-frequency components of the data.
    }
    \label{fig:conceptdiagram}
\end{figure*}

(ii) \textbf{Feature reusability under heterogeneous data}:  MRI is integral and multi-parametric with heterogeneous contrasts that offer multimodal (diverse and complementary) information for radiological decision-making, enabling specialized applications in neurodiagnosis, spine, cardiac, musculoskeletal, and soft tissue, extending to chronic and degenerative diseases like cancer.
Training bespoke models specific to each modality or learning a common model by combining samples from different modalities might not be viable solutions for high-quality imaging workflows as (a) modality-specific models restrict the AM to capturing the local contextual variations within a specific domain and (b) they necessitate retraining under distribution shifts in MRI data \cite{univusmri}. Joint-trained models that combine  data from multimodal MRI datasets learn an abstract mode-invariant representation \cite{mcihypernet} where the AM might not adequately capture the subtle differences associated with different frequency components of the multimodal data. A static set of features might not adequately capture the information about the different levels of transferable features of multiple domains to generalize to drifted domains \cite{dff}.

These challenges motivate the need for a discriminative multi-spectral AM that can propagate the context-aware high-frequency details for high-quality image reconstruction, simultaneously capture different transferable or reusable features from multiple seen to deviated unseen domains to offer high practical value for the healthcare industry and researchers \cite{neuralizer}. We consider the problem of implicit local and global feature modeling from frequency perspectives in attention-based deep neural networks for a unified MRI reconstruction, generalizable across widely varying acquisition settings. 

Neuromodulation, an emerging direction in brain-inspired deep learning \cite{braininspired, neuralbasis} is an attentional modulation technique \cite{neuralbasis,neuromod_nn} that can continuously tune the neuron’s response in different contexts, generally in response to an external input signal. In biological neural systems, neuromodulation regulates many nervous system properties critical to the adaptive control of continuous behaviors. 
Inspired by these merits, we propose an architecture with neuromodulation-based AM to endow the deep neural network for MR image reconstruction with an adaptive learning capability similar to biological neural networks. The proposed architecture comprises two sub-networks - a dynamic spectral filtering CNN and a high-pass kernel generation transformer. The spectral filtering CNN provides dynamic frequency-domain attention information that recalibrates the features to enhance the transferable frequency components from seen to unseen MRI data across multiple MRI modalities and varying under-sampling degradation levels in the image reconstruction. The dynamic high-pass kernel generation transformer learns the relationship between distant regions, focusing on high-frequency details for improved reconstruction.
The concept diagram comparing the previous neuromodulation techniques with the proposed method is shown in Figure \ref{fig:conceptdiagram}.
The proposed neuromodulation method performs instance-specific AM to provide local contextual and global mode-specific features of multimodal MRI data with frequency perspectives. Our experiments assert that the proposed method applies to self-supervised learning, on-the-fly adaptation, and generative models like diffusion models.
Our contributions are:

1. We propose SHFormer, an AM-driven neural network architecture consisting of (i) a deep spectral filtering CNN, and (ii) a dynamic high pass kernel generation transformer for adaptive MRI reconstruction.

2. The proposed hybrid AM performs neuromodulation in the spectral and spatial domains. The spectral AM (i) learns discriminative mode-specific attention weights in the frequency domain, and (ii) selectively utilizes the learned meta-knowledge to adapt to the target data. The instance-specific spatial AM learns high-frequency details for higher reconstruction fidelity.

3. The proposed network uses dynamic weight learning auxiliary networks that are conditioned on the convolution and self-attention features and predict implicit sample-specific filters, incorporating multiple contextual information in a single model and expanding the feature space to multimodal MRI contrasts and varying acceleration factors.

4. The proposed network considers two perspectives crucial to clinical deployment: (i) closed-set generalization via a continuously varying implicit feature representation to unseen undersampling levels and open-set generalization to multi-contrast MRI data, and (ii) propagating the high-frequency signals for better recovery of structures.

5. We have extensively evaluated our proposed network from various perspectives in MRI reconstruction, namely (i) comparative studies across (a) varying learning modes - supervised and self-supervised, and generative model setting, (b) various task-based and instance-based adaptive methods, (ii) visual interpretation of the learned features, and (iv) ablative study. 

6. The proposed network exhibits promising results with best improvement margins of (i) 0.3 to 0.6 dB PSNR and 0.006 to 0.01 in SSIM in supervised and self-supervised learning, outperforms the baseline diffusion model without spectral and dynamic high-pass filtering, (ii) $\sim$1 dB in PSNR and $\sim$0.01 in SSIM for closed and open-set generalization experiments.

\begin{figure*}[t!]

    \centering
    \includegraphics[width=0.8\linewidth]{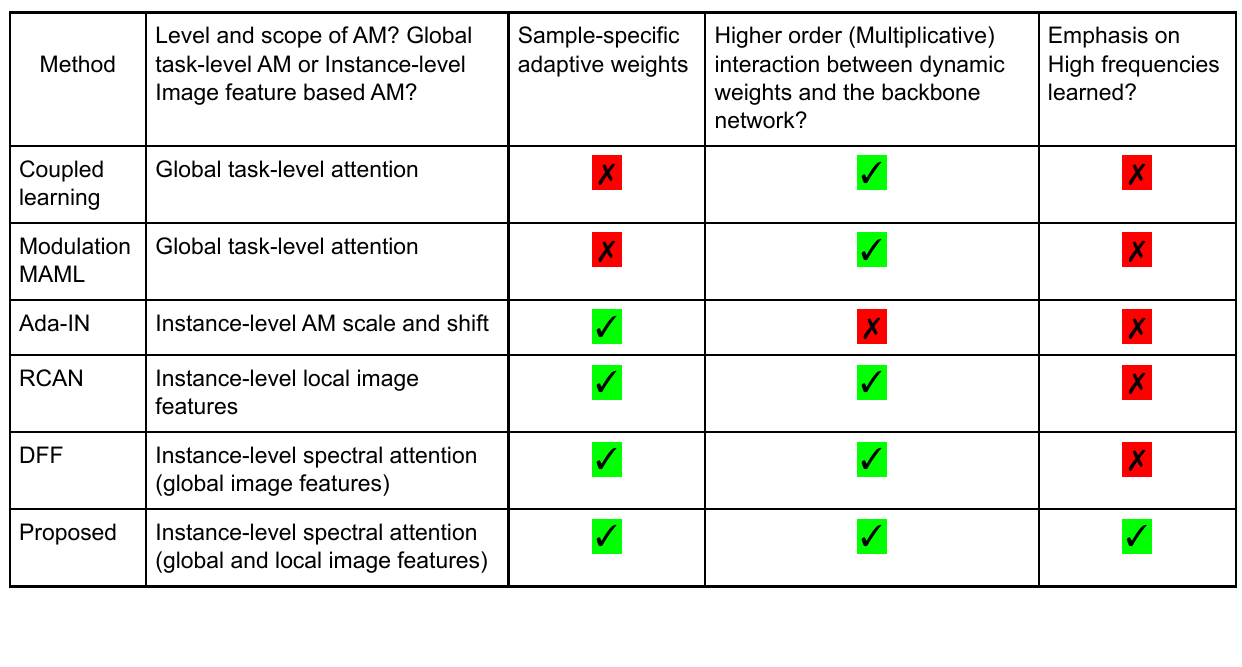}

    \caption{A tabulated summary of various neuromodulation-based AMs (coupled learning \cite{mcihypernet}, modulation MAML \cite{kmmaml}, adaptive instance normalization (Ada-IN) \cite{univusmri}, channel attention (RCAN) \cite{rcan}, and deep frequency filtering (DFF) \cite{dff}) and the proposed AM approach for MRI reconstruction comparing various aspects -  the scope of attention and adaptability, the type of interactions, and frequency perspectives
    }
    \label{tableconceptdiagram}
\end{figure*}
\FloatBarrier
\section{Related Work}
Adaptive learning approaches from the perspectives of AM can be broadly classified as task-level and instance-level (or sample-specific) methods based on the level of abstraction at which important information is captured. 

\subsection{Task-level Adaptive Methods}
Task-level adaptive methods support multiple parameterized image processing operators in a single network training, considering each operator as an imaging task. 
Gradient-based or Model agnostic meta-learning (MAML) adopts two-level optimization of the network to learn task-specific weights using support and query data of tasks, yet relies on a single meta-initialization \cite{maml} that may need test-time training to adapt to the target data \cite{ontheflytta, tent}. Architecture-based methods employ an adaptive weight learning network, called hypernetwork \cite{hypernetworks}, as an auxiliary learning network to learn task-specific weights. There are a few attempts that integrate weight learning networks with a backbone imaging task-oriented network for representing multiple imaging tasks in a single network. The decouple learning framework \cite{gnldecouple} uses a weight learning network driven by the imaging operator parameter (ex., scale factor in image super-resolution) to predict all the weights of the backbone network. The controllable image processing networks \cite{cfs-net,controllable} use a weight learning network to predict parameters that adjust the functionality of the backbone network to support multiple imaging tasks. The task-aware modulation network \cite{mmaml} uses weight learning networks that predict parameters that scale the backbone network layer weights for few-shot learning and image classification tasks. Different from these approaches, the weight learning network in our method does not need explicit context embedding and predicts adaptive weights based on implicit features. Moreover, our method provides global attention information at the task level via sample-specific spectral filtering and local attention information in the image features.  

\subsection{Instance-level Adaptive Methods}
Instance-level adaptive methods 
adjust attention weights for each sample, enhancing context-aware processing by selectively concentrating on important information within features. The residual channel attention (RCAN) \cite{rcan} uses a set of global attention descriptors for channels via global average pooling 
to calculate the channel weights.  The self-attention mechanism in vision transformers \cite{vit, swin, restomer} computes attention weights based on the similarity between the query and key vectors, focusing more on relevant patches in the input image. The adaptive instance normalization (Ada-IN) \cite{taskswitch} is another variant of sample-specific AM to adjust the mean and standard deviation of feature maps under heterogeneous data. Unlike image domain AM, spectral AM offers a frequency-based approach, enhancing generalization under distribution shifts from the training data domain. The frequency channel attention network, FcaNet \cite{fcanet}, extends RCAN by using discrete cosine transform (DCT) to compute global attention weights to channels, assigning a fixed frequency component to each group of channels. The matrix multi-spectral channel attention (MMCA) \cite{mmca} assigns the topmost DCT coefficient from multiple frequency components and extracts high-frequencies from convolution features. 
The deep frequency filtering (DFF) learns a discrete Fourier transform domain spatial attention map applied to the DFT of the channels. These methods abstract out the high-frequency information via max and mean pooling.
Our approach uses (i) DCT-domain spectral AM along spatial and channel dimensions comprehensively without pooling, utilizing the dominant spectral components for enhanced transferability of features to heterogeneous data, and (ii)  extracts high-frequency information of self-attention features enabling global and local context.


\subsection{Adaptive Methods in MRI}
Several adaptive learning methods for MRI exist in the literature. Deep neural networks like MAC-ReconNet \cite{mac} and MCI-HyperNet \cite{mcihypernet} perform acquisition context-adaptive MRI reconstruction by considering combinations of acceleration factor, undersampling pattern, and the anatomy under study as an acquisition context or a task using dynamic weight learning auxiliary network. The MAC-ReconNet predicts the task-specific weights of a deep cascaded CNN reconstruction network.  MCI-HyperNet infuses context-specific weights into a UNet-like backbone reconstruction network by convolving the context-specific weights with the backbone network features. Model-agnostic meta-learning (MAML) \cite{maml,metal_survey} methods like kernel modulation-based meta-learning (KM-MAML) \cite{kmmaml} and Curriculum MAML (CMAML) \cite{cmaml} learn context-specific weights via two-levels optimization for multi-contrast MRI data and multiple artifact-affected MRI data, respectively. The universal under-sampled MRI reconstruction \cite{univusmri} employs Ada-IN and knowledge distillation (KD) to combine multiple anatomies, necessitating anatomy-specific training.
Instance-level adaptive methods are MICCAN \cite{miccan}, a channel AM-based method for MRI reconstruction, and on-the-fly test-time adaptation for MRI segmentation \cite{ontheflytta}. All these methods incline towards low frequencies as given by the F-principle \cite{TrainbehaviourDNNFreq}. The proposed method ensures (i) adaptive learning and generalization to deviated multimodal data via spectral filtering, and (ii) high-frequency information propagation through the network via dynamic high-pass kernel generation (as summarized in the table in Figure \ref{tableconceptdiagram}).

\begin{figure*}[t!]
    \centering
    \includegraphics[width=\linewidth]{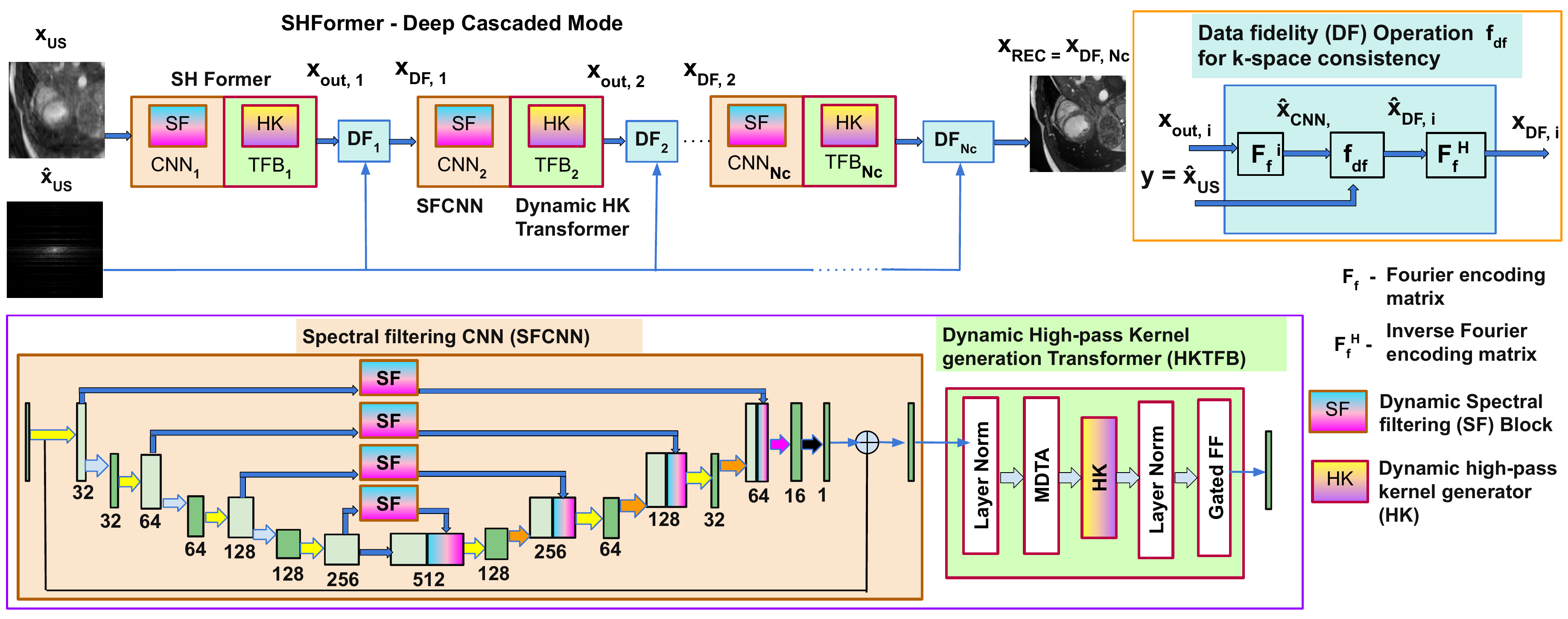}
    \caption{(Top Left) Architecture block diagram of SHFormer in the deep cascaded mode. (Bottom) Detailed network structure SHFormer. The spectral filtering module in the CNN extracts a spatial attention map that captures mode-specific reusable features of multimodal data. The dynamic kernel generation transformer emphasizes high-frequency components to be passed to the normalization layer for improved structural information flow. (Top right) k-space data fidelity operation
    }
    \label{fig:archblockdiagram}
\end{figure*}

\begin{figure*}[t!]
    \centering
    \includegraphics[width=\linewidth]{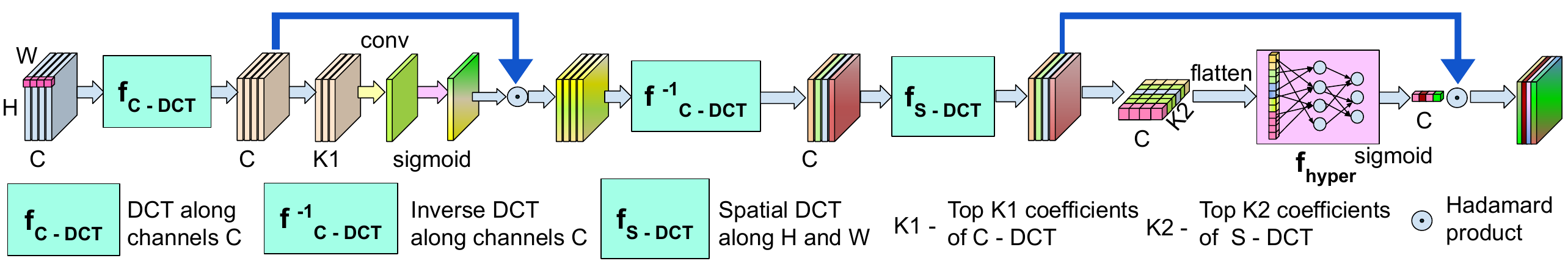}
    \caption{Spectral filtering CNN (SFCNN) structure: The SFCNN consists of two DCT-based spectral AM stages operating on the encoder features of the CNN. The first stage learns a spatial frequency domain attention map by applying forward and inverse DCT along channel direction ($f_{C - DCT}$ and $f^{-1}_{C-DCT}$, respectively) at each spatial location of the CNN features. The second stage learns spectral DCT-based attention values by applying 2D DCT ($f_{S - DCT}$) of each channel spatially in the image domain.
    }
    \label{fig:sfcnn}
\end{figure*}
\FloatBarrier
\section{Method}

\subsection{Preliminary and Problem Formulation}
The data acquisition forward model of the MR image reconstruction problem \cite{dagan} can be formulated as a linear system as follows:
\begin{align}
 \label{Forward_model}
 \mathcal{A}x+\epsilon = y
\end{align}

Here, $x\in \mathbb{C}^N$ denotes the desired image, $y$ $\in$ $\mathbb{C}^{M'}$ is the under-sampled (US) measurement from the MRI scanner, $\epsilon \in \mathbb{C}^{M'}$ is the noise. 
The under-sampled image reconstruction is ill-posed as the problem is under-determined ($M' << N$), and the operator $\mathcal{A}$ is ill-conditioned. The under-sampled (US) or zero-filled (ZF) image is given by $x_{US} = F_{US}^{H}y$ where $F_{US} = M \odot \mathcal{F}$ is the under-sampled Fourier encoding matrix representing the forward model $\mathcal{A}$. 
Here, $M$ is a 2-D under-sampling mask, $\odot$ represents Hadamard product, and $\mathcal{F}$ is 2-D Fourier transform, respectively. 
The ZF image, $x_{US}$, is an aliased image due to sub-Nyquist sampling. The reconstruction of the under-sampled image is achieved by introducing an apriori knowledge of $x$ into the unconstrained optimization \cite{dagan} and is given as:

\begin{align}
 \label{Optimization formulation}
 \underset{x} {\operatorname{min}} \quad ||\mathcal{A}x - y||_{2}^2 + \mathcal{R}(x)
\end{align}

where, $||\mathcal{A}x - y||_{2}^2$ is the k-space  data fidelity term \cite{dc_cnn} and $\mathcal{R}$ is a regularization term.

\subsubsection{Supervised learning}

Deep learning-based MRI reconstruction involves training a deep learning (DL) model by optimizing the weights of a neural network to minimize the average loss of all observed data samples and can be formulated as:
\begin{align}
 \label{Joint-training formulation}
 \theta^{*} = \underset{\theta} {\operatorname{argmin}}  \displaystyle \mathop{\mathbb{E}}_{(x_{US},x_{FS})\in \bigcup\limits_{i}^{} \mathcal{D}_{i}}[||x_{FS} - f(x_{US};\theta) ||_{2}^2]
\end{align}

Here, $\mathcal{D}_{i}$ represents the dataset of the $i^{th}$ configuration of MRI contrast, under-sampling mask type, and acceleration factor, consisting of under-sampled and fully sampled (FS) (ground truth or GT) image pairs, $(x_{US}, x_{FS})$. Here, $f$ is the DL model parameterized by $\theta$ with k-space data fidelity (as shown in Figure \ref{fig:archblockdiagram}). 

\subsubsection{Self-Supervised learning }

 Following the physics-driven self-supervised learning approach \cite{yaman2020self}, we randomly partition the k-space, $y$, into two disjoint sets $y_1$ and $y_2$  as follows.
 \begin{align} 
 y_1 = M_1 \odot y \quad and \quad
 y_2 = M_2 \odot y
 \end{align}
where $M_1$ and $M_2$ are the two disjoint masks that partition the k-space $y$. One partition $y_1$ is fed as input to the neural network while the other partition $y_2$ is taken as the GT measurement set for optimizing the parameters $\theta$ of the neural network $f(y_1;\theta)$. The loss function $L$ is defined as,  
\begin{equation}
    L(y_1,y_2) =  || M_2 \odot \mathcal{F}(f(y_1;\theta)) - y_2||_{1} 
\end{equation}
This self-supervised approach optimizes the parameters of the neural network, eliminating the need for GT images. 

\begin{figure*}[t!]
    \centering
    \includegraphics[width=0.8\linewidth]{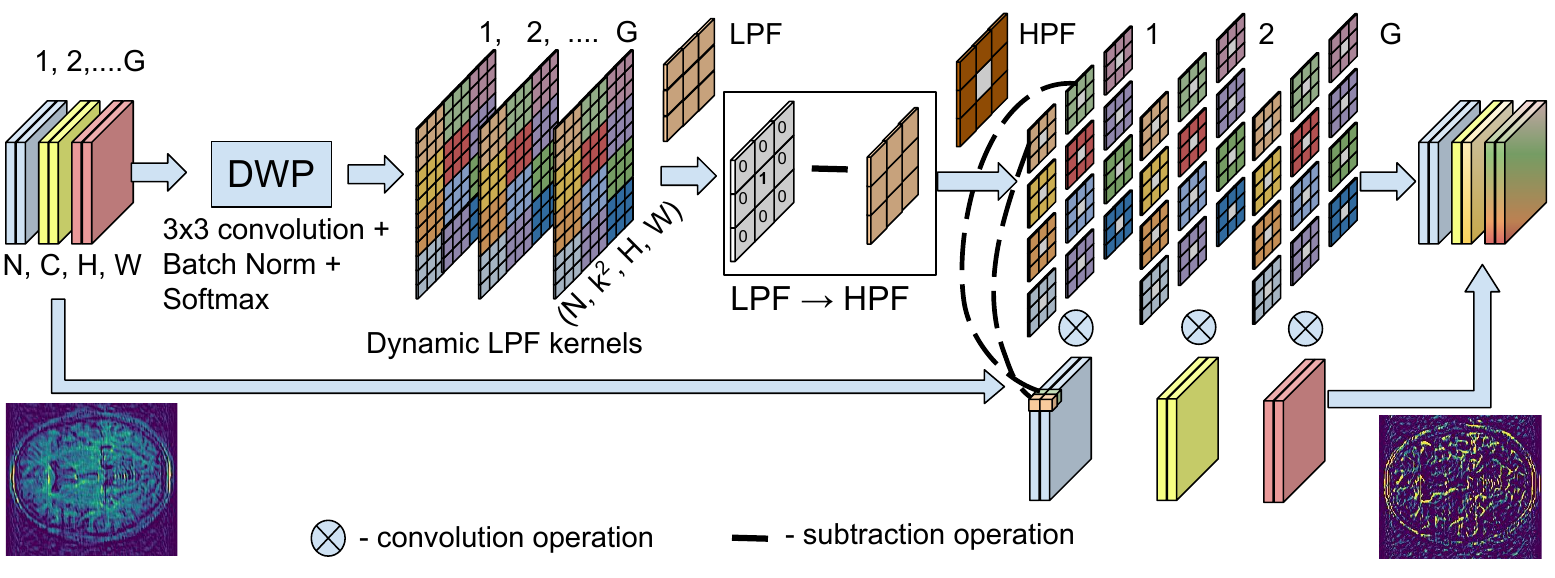}
    \caption{Dynamic High-pass kernel generation (HK) block consists of a DWP network that predicts instance-specific low-pass filter kernels along each channel group for each spatial location of the features. The low-pass filter kernels are converted to high-pass filter kernels and applied to the input channels to extract details.
    }
    \label{fig:hpf}
\end{figure*}

\subsubsection{Diffusion model-based learning}

Motivated by the superior performance of the denoising diffusion probabilistic model \cite{ddpm} in various biomedical imaging tasks and the role of self-attention in diffusion models in learning visual concepts \cite{conceptrasediff, palette}, we propose a dynamic spectral and high-pass filtering-induced measurement-constrained denoising diffusion probabilistic model, MC-DDPM (SH) as a variant of the probabilistic model, MC-DDPM \cite{mcddpm}.
Diffusion models involve forward and reverse processes (More details in the supplementary material). The forward process gradually adds noise to the input data until it is transformed into pure Gaussian noise. In the reverse process, a denoising network consisting of dynamic spectral filtering in the skip connections and high-pass kernel generation in the self-attention blocks, predicts the noise at each time step. In the proposed measurement-constrained diffusion model, the reverse process is in the measurement domain, where the noise is only added at non-sampled positions in the k-space, which are zero-valued. We define this mask as $M^c = I - M $, the complement of the undersampling mask $M$.
(the superscript c means complement) and $y_{M^c} = M^cAx$, which represents the non-sampled k-space
measurements. 
The sampling process, given by $p_{\theta}(y_{M^c,0} | M^c,y_M)$ is the reverse process of the diffusion model. Training of MC-DDPM (SH) is performed by optimizing the variational bound on the negative log-likelihood of $p_{\theta}(y_{M^c,0} | M^c,y_M)$ and this loss function \cite{mcddpm} is given as:

\begin{equation}
\begin{split}
\mathcal{L} =   E_{y_{M^c,0},t,\epsilon} ||\epsilon - \epsilon_{\theta} (\bar{\alpha_t}y_{M^c,0} + \bar{\beta_t}\epsilon,t,M^c,y_M ) ||^2_{2}, \\ \epsilon \sim \mathcal{N}(0,M^c)
\end{split}
\end{equation}
Here $\alpha_t := 1 - \beta_t$ and $\beta_t$ gives the variance schedule of the forward process at an arbitrary time step $t$,  $\bar{\alpha}_t := \prod_{i=1}^{t}\alpha_t$,  $\bar{\beta}_t = \sum_{i=1}^{t}\frac{\bar{\alpha}^2_{t}}{\bar{\alpha}^2_{i}}\beta^2_{t}$ (with $\bar{\alpha}_0 = 1$ and $\bar{\beta}_0 = 0$), and $\epsilon \sim \mathcal{N}(0,M^c)$.

\subsection{Proposed Method}
The architecture of SHFormer is a cascaded structure with alternating reconstruction network and k-space data fidelity units. The reconstruction network consisting of spectral filtering encoder-decoder-based CNN (SFCNN), and a dynamic high-pass kernel generation transformer block (HKTFB) (Figure \ref{fig:archblockdiagram}) is illustrated in detail as follows.

\subsubsection{Spectral Filtering CNN}
The spectral filtering CNN consists of a dynamic spectral filtering module and a channel-wise spectral attention module along the skip connections of the encoder-decoder-based CNN (Figure \ref{fig:sfcnn}). 
The spectral filtering module learns an instance-adaptive spatial mask to dynamically modulate different frequency components of images from multimodal data
during training for learning generalizable mode-invariant and mode-specific decoder features. 
Specifically, we obtain spectral components using DCT \cite{fcanet} for the following reasons: (i) DCT 
exhibits energy compaction as a significant portion of the total signal energy is concentrated in the top few DCT coefficients.
(ii) DCT is computationally efficient and real-valued. 
Let $X_F$ be the input feature of size $(N, C, H, W)$ where N is the batch size, $C$, $H$, and $W$ are the dimensions along channels, height, and width of the feature, respectively.
The first step in spectral filtering is the DCT
applied along the channel direction $C$ considering the channel-wise feature values at each spatial location as a $1D$ sequence $x[n]$ where $n = 0, 2, .. C-1$ and is given as
\begin{equation} \label{cdct}
f_{C-DCT}[k] = a(k) \sum_{n=0}^{C - 1} x[n] cos \left(\frac{\pi(2n+1)k}{2C}\right)
\end{equation}
Here, the index, $k = 0, 1,.., C-1$
\[ a[k] =
    \begin{cases}
        \sqrt{\frac{1}{C}} & k = 0 \\
        \sqrt{\frac{2}{C}} & 1 \leq k \leq C-1 \\
    \end{cases}
\]
The top $K1$ coefficients of the $f_{C-DCT}$ output are taken, followed by the spatial frequency domain attention mechanism consisting of a convolution layer that outputs the attention map, followed by sigmoid activation.
\begin{equation}
X_{SF,DCT} = f_{C-DCT}(X_F)\; \odot \; \sigma (conv(f_{C-DCT}(X_F))) 
\end{equation}

Here $\odot$ denotes the element-wise multiplication
of the spatial attention mask with DCT-domain features of $X_F$, $conv$ indicates the convolution layer and $\sigma$ denotes sigmoid activation of the AM.
The final attention map is used to reweigh the DCT maps spatially. The inverse DCT $f^{-1}_{C-DCT}$ is taken along the channel dimension to transform the features to the intensity domain.
\begin{equation}
f^{-1}_{C-DCT}[n] = \sum_{u=0}^{C - 1} a[u] X_{SF,DCT}[u] cos\left(\frac{\pi(2n+1)u}{2C}\right) 
\end{equation}

\begin{equation}
X_{SF} = f^{-1}_{C-DCT}(X_{SF,DCT}) 
\end{equation}
Here $a[u]$ is defined in Equation \ref{cdct}. The spectral filtering is followed by a channel-wise spectral attention mechanism where 2D DCT is applied along the spatial dimensions $H$ and $W$ for each channel. The 2D DCT is given by the standard formulation as follows.

\begin{equation}
\begin{split}
 f_{S - DCT}[h,w] = \sum_{p=0}^{H - 1} \sum_{q=0}^{W - 1} X_{SF}[p,q] \; cos\left(\pi \frac{h}{H}(p+\frac{1}{2})\right) \\ cos\left(\pi \frac{w}{W}(q+\frac{1}{2})\right)
\end{split}
\end{equation}

The top $K2$ coefficients of $f_{S - DCT}$ along the spatial dimension yield an attention vector of length $K2$ for each of the $C$ channels. This gives $K2*C$ outputs fed to a fully connected hypernetwork $f_{hyper}$ with a sigmoid activation in the output layer that produces $C$ spectral attention weights. These weights recalibrate the spectral filtering output features $X_{SF}$ to give the final output features $X_{F, out}$. 

\begin{equation}
X_{S,DCT} = f_{S - DCT}(X_{SF}) 
\end{equation}

\begin{equation}
X_{F, out} = f_{hyper}(X_{S,DCT})\; \odot \; X_{SF} 
\end{equation}

\subsubsection{Dynamic High-pass Kernel Generation Transformer} 
The HKTFB (shown in Figure \ref{fig:archblockdiagram}) consists of a layer normalization module, self-attention blocks, the instance-specific dynamic high-pass filtering module, and the output layer normalization module. 
Unlike the dynamic HPF module in MMCA \cite{mmca}, which operates on convolution features, the proposed transformer learns high-frequency features from the self-attention features. 
The input to the HKTFB is $X \in R^{C\times H \times W}$, where $C$ is the number of channels, $H$ is the height and $W$ is the width of the image.

The queries $Q$, keys $K$, and values $V$ of the transformer are computed by applying the linear transformation as in Equation \ref{qkv_compute}.  
The transformations consist of $1\times1$ and $3\times3$ depth-wise convolutions to capture cross-channel and channel-wise spatial context using learnable weight matrices $W_p^{(.)}$ and $W_d^{(.)}$, respectively.
\begin{equation} 
\label{qkv_compute}
\begin{split}
Q=W_d^QW_p^QX \quad K=W_d^KW_p^KX \quad
V=W_d^VW_p^VX
\end{split}
\end{equation}
To facilitate attention computation, Q, K, and V are resized as $\hat{Q}, \hat{K},\hat{V} \in R^{C \times HW}$. The attention map, $A \in R^{C \times C}$, is generated via the dot product interaction between the query and key value followed by the softmax function as follows.
\begin{equation} 
\label{attn_compute}
\begin{split}
A = Attention(\hat{Q}, \hat{K},\hat{V}) = softmax(\frac{\hat{Q}\hat{K}^T}{\alpha}) 
\end{split}
\end{equation}
Here $\alpha$ is a learnable parameter that controls the saturation of the softmax function. The attention-weighted features are given as,
\begin{equation} 
\label{attnout_compute}
\hat{X} = W_p A \hat{V} + X
\end{equation}
The attention-weighted features $\hat{X} \in R^{C, H, W}$ are passed to an instance-specific high-pass filtering module (Figure \ref{fig:hpf}) similar to MMCA \cite{mmca}. Unlike the dynamic HPF module in MMCA, which operates on convolution features, the proposed transformer learns high-frequency features from the self-attention features.
The high-frequency features are passed to the feed-forward network with a gating mechanism with depth-wise convolutions of the transformer \cite{restomer}, which helps to learn high-frequency information for better structure recovery. 

The input features are converted to $G$ groups and are passed to a convolutional hypernetwork (denoted as $Conv$) followed by batch normalization ($BN$). 
The hypernetwork generates unique kernels $w_{hyper} \in R^ {G \times K\times K, H, W}$, where $K$ is the kernel size.
These kernels are resized and passed to a softmax activation to represent low-pass filters. The LPF, $LPF(i,k,h,w)$, for each group $i$ at each spatial location $(h,w)$ is a $K\times K$ kernel that sums up to one.
\begin{equation} 
\label{hpfilter_compute}
\begin{split}
LPF(i,k,h,w) = softmax (w_{hyper}(i,k,h,w)), \\ 
\forall\  i = 1, ..., G, \quad
\forall\  k = 1, ..., K^2, \quad
\forall\ h = 1, ..., H, \\
\forall\ w = 1, ..., W  \\
\end{split}
\end{equation}

The dynamic high-pass filter, $DynHPF$ is obtained by inverting the low-pass filter (such that each kernel sums down to zero) and applied at each spatial location in each group of input features.
\begin{equation}\label{eq_lpf_to_hpf}
DynHPF(i,k,h,w) =
    \begin{cases}
        LPF(i,k,h,w) -1 & k = \lfloor\frac{K^2}{2}\rfloor  \\
        LPF(i,k,h,w) & otherwise \\
    \end{cases}
\end{equation}
Here $\lfloor\frac{K^2}{2}\rfloor$ denotes the middle element index of the kernel. The final attention equation is given as,

\begin{equation} 
\label{finalattn}
\hat{X}_{out} = DynHPF(W_p\ Attention(\hat{Q}, \hat{K},\hat{V})\ \hat{V}) + X
\end{equation}


\begin{table*}[]
\centering
\scriptsize
\caption{Quantitative Comparison of SHFormer with other single-coil MRI reconstruction architectures on large-scale datasets (1) ACDC cardiac challenge datasets and (2) complex-valued realistic clinical knee MRI (FastMRI) Knee dataset }
\label{tab:sc_cardiac_knee}
\setlength{\tabcolsep}{5.0pt}
\renewcommand{\arraystretch}{1.3}
\begin{tabular}{@{}ccccccccc@{}}
\toprule
\multirow{3}{*}{\textbf{Method}} & \multicolumn{4}{c}{\textbf{Cardiac dataset}}                                                                    & \multicolumn{4}{c}{\textbf{Knee dataset}}                                                                       \\ \cmidrule(l){2-5} \cmidrule(l){6-9}
                                 & \multicolumn{2}{c}{\textbf{4x}}                        & \multicolumn{2}{c}{\textbf{5x}}                        & \multicolumn{2}{c}{\textbf{4x}}                        & \multicolumn{2}{c}{\textbf{8x}}  
                                 \\ \cmidrule(l){2-3} \cmidrule(l){4-5}  \cmidrule(l){6-7} \cmidrule(l){8-9} 
                                 & \textbf{PSNR}             & \textbf{SSIM}              & \textbf{PSNR}             & \textbf{SSIM}              & \textbf{PSNR}             & \textbf{SSIM}              & \textbf{PSNR}             & \textbf{SSIM}              \\ \midrule
ZF                               & 24.27 $\pm$ 3.10          & 0.6996 $\pm$ 0.08          & 23.82 $\pm$ 3.11          & 0.6742 $\pm$ 0.08          & 28.40 $\pm$ 4.39          & 0.7504 $\pm$ 0.17          & 27.82 $\pm$ 4.38          & 0.7270 $\pm$ 0.16          \\
DAGAN \cite{dagan}                           & 28.52 $\pm$ 2.71          & 0.8410 $\pm$ 0.04          & 28.02 $\pm$ 2.80          & 0.8248 $\pm$ 0.05          & 30.28 $\pm$ 4.34          & 0.8284 $\pm$ 0.15          & 29.62 $\pm$ 4.15          & 0.8138 $\pm$ 0.14          \\
DC-CNN \cite{dc_cnn}                          & 32.75 $\pm$ 3.28          & 0.9195 $\pm$ 0.04          & 31.75 $\pm$ 3.40          & 0.9054 $\pm$ 0.04          & 30.90 $\pm$ 4.47          & 0.8439 $\pm$ 0.16          & 30.38 $\pm$ 4.00          & 0.8418 $\pm$ 0.13          \\
DC-UNet \cite{dc_unet}                         & 33.17 $\pm$ 3.60          & 0.9276 $\pm$ 0.04          & 32.55 $\pm$ 3.71          & 0.9189 $\pm$ 0.04          & 31.17 $\pm$ 4.53          & 0.8542 $\pm$ 0.15          & 30.75 $\pm$ 4.03          & 0.8514 $\pm$ 0.13          \\
DC-DEN \cite{dc-ensemble}                          & 33.22 $\pm$ 3.46          & 0.9249 $\pm$ 0.04            & 32.30 $\pm$ 3.57          & 0.9126 $\pm$ 0.04          & 30.43 $\pm$ 7.80          & 0.8370 $\pm$ 0.16          & 29.95 $\pm$ 3.90          & 0.8255 $\pm$ 0.13          \\
DC-RDN \cite{recursive_dilated}                          & 32.95 $\pm$ 3.40          & 0.9233 $\pm$ 0.04          & 32.09 $\pm$ 3.57          & 0.9115 $\pm$ 0.04          & 29.92 $\pm$ 4.28          & 0.8119 $\pm$ 0.16          & 29.53 $\pm$ 3.94          & 0.8081 $\pm$ 0.14          \\
MICCAN \cite{miccan}                          & 33.34 $\pm$ 3.51          & 0.9287 $\pm$ 0.04          & 32.60 $\pm$ 3.66          & 0.9192 $\pm$ 0.04          & 31.20 $\pm$ 4.55          & 0.8547 $\pm$ 0.15          & 30.80 $\pm$ 4.06          & 0.8532 $\pm$ 0.13          \\
DC-Hybrid \cite{hybrid}                       & 32.07 $\pm$ 3.35          & 0.9097 $\pm$ 0.04          & 30.93 $\pm$ 3.51          & 0.8930 $\pm$ 0.05          & 31.00 $\pm$ 4.46          & 0.8511 $\pm$ 0.15          & 30.32 $\pm$ 3.95          & 0.8423 $\pm$ 0.13          \\
OUCR \cite{
}                            & 33.29 $\pm$ 3.45          & 0.9272 $\pm$ 0.04          & 32.29 $\pm$ 3.52          & 0.9139 $\pm$ 0.04          & 30.69 $\pm$ 4.46          & 0.8365 $\pm$ 0.16          & 30.08 $\pm$ 3.98          & 0.8296 $\pm$ 0.14          \\
SWIN \cite{swin}                            & 30.15 $\pm$ 2.96          & 0.8686 $\pm$ 0.04          & 29.24 $\pm$ 2.98          & 0.8479 $\pm$ 0.05          & 31.02 $\pm$ 4.45          & 0.8513 $\pm$ 0.15          & 30.38 $\pm$ 4.00          & 0.8418 $\pm$ 0.13          \\
SFT-KD-Recon \cite{sftkdrecon}                    & 32.03 $\pm$ 3.19          & 0.9070 $\pm$ 0.04          & 30.93 $\pm$ 3.28          & 0.8884 $\pm$ 0.05          & 30.15 $\pm$ 3.95          & 0.8339 $\pm$ 0.13          & 29.21 $\pm$ 4.27          & 0.7949 $\pm$ 0.14          \\
SHFormer                         & \textbf{33.80 $\pm$ 3.56} & \textbf{0.9342 $\pm$ 0.03} & \textbf{32.84 $\pm$ 3.66} & \textbf{0.9222 $\pm$ 0.04} & \textbf{31.24 $\pm$ 4.55} & \textbf{0.8568 $\pm$ 0.15} & \textbf{30.85 $\pm$ 4.10} & \textbf{0.8545 $\pm$ 0.13} \\ \bottomrule
\end{tabular}
\end{table*}

\FloatBarrier
\section{Experiments}

We broadly categorize our experiments as follows:  1. Comparative study-based to assess the performance of the proposed approach against other reconstruction methods based on aspects such as (i) assessment on a large-scale collection of clinical and complex-valued realistic datasets from multiple scanner sites (ii) diffusion model-based evaluation in a generative setting, and (iii) Self-supervised learning 2. Generalization-based, such as (i) closed-set domain generalization and (ii) Open-set domain generalization 3. Ablative study-based, and 4. Visualization-based to interpret the learned features.

\subsection{Dataset}
We have evaluated our model using the following datasets.
For comparison against other CNN-based, GAN-based, KD-based and transformer-based models,  we have used the following datasets.

1. \textbf{Automated Cardiac Diagnosis Challenge (ACDC) cardiac MRI} dataset \cite{acdc_dataset}, which consists of 150 and 50 patient records with 1841 and 1076 slices for training and validation, respectively. The 2D slices are extracted and cropped to 150 $\times$ 150. 

2. \textbf{FastMRI}  \cite{fastmri} knee dataset contains 7040 training, 3317 validation, and 3903 testing slices of 200, 99 and 108 volumes, respectively, of coronal proton-density without (PD) and with fat suppression (PDFS).

For self-supervised multi-coil MRI reconstruction, we have used the following dataset.

3. The \textbf{Multicoil knee dataset} \cite{variational} consists of three protocols: coronal proton-density (PD), coronal fat-saturated PD (PDFS), and axial fat-saturated T2. The data was acquired through a 15-channel multi-coil setting for 20 subjects. Each 3D volume has 40 slices of 640 $\times$ 368 resolution complex-valued data and the corresponding sensitivity maps. The center 19 slices were considered for our evaluation. The dataset was partitioned into 10 volumes containing 190 slices each for training and 10 volumes with 190 slices each for validation.

For the closed-set generalization experiment, we have used the ACDC Cardiac dataset. For the open-set generalization experiment, we have used multimodal MRI brain datasets (MRBrainS and IXI) consisting of four multi-contrast images, as illustrated below.

4. \textbf{MRBrainS} dataset \cite{mrbrains_dataset}: We consider 336 slices of T1 and FLAIR contrasts from 7 volumes of T1 and FLAIR (fluid-attenuated inversion recovery) with T1: TR = 7.9 ms, TE = 4.5 ms and FLAIR: TR = 11s and TE = 125 ms acquired on a Philips scanner. The repetition and echo times, TR, and TE denote the contrast-specific MRI settings. All the images have a size of $240\times240$.

For diffusion model-based comparative studies, we have used the MRBrains dataset.

5. \textbf{IXI\footnote{https://brain-development.org/ixi-dataset/}} brain dataset: We consider 1400 axial brain slices of T2 and Proton density (PD) weighted contrasts acquired from 14 volumes with T2: TR = 5725 ms, TE = 100 ms, and PD: TR = 5725 ms, TE = 8 ms acquired on a Philips scanner. The images are pre-processed by cropping to a size of $240\times240$.

6. \textbf{Calgary} dataset: The human brain dataset, released by \cite{article2}, consisted of 45 T1 volumes. The center 110 slices of each volume were considered for the experiments, resulting in 25 volumes or 2750 slices for training and 10 volumes or 1100 slices for validation. The data was acquired through a 12-channel multi-coil setting and combined along the coils' dimensions to simulate a single coil acquisition. The complex-valued MRI slices were of size $256\times256$.

\begin{figure*}[t!]
    \centering
    \includegraphics[width=\linewidth]{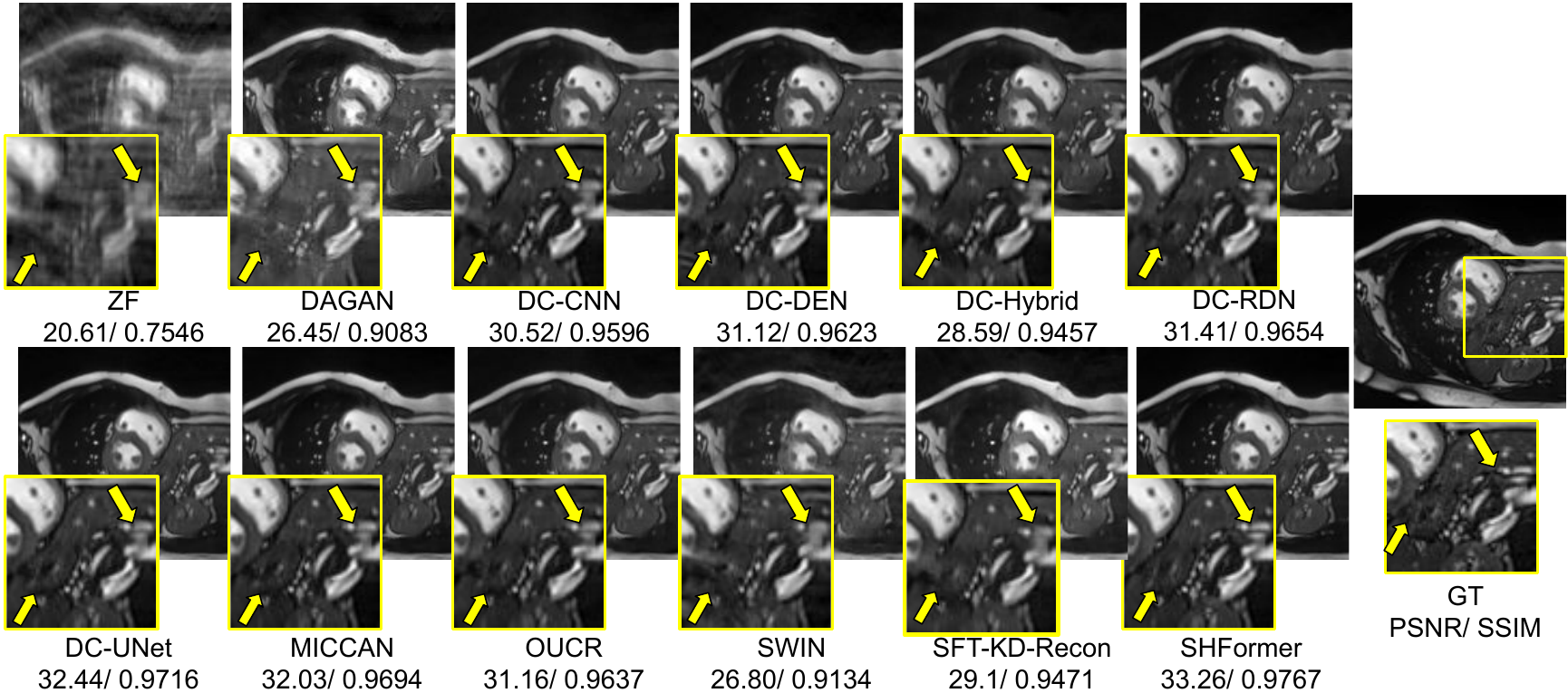}
    \caption{Qualitative comparison of the predictions of SHFormer with other CNN-based, GAN-based, transformer-based, and KD-based methods for 5x cardiac MRI reconstruction.  The yellow arrows show that the visual quality of the proposed model is much better for SHFormer than other reconstruction methods.
    }
    \label{fig:sc_cardiac}
\end{figure*}

\subsection{Implementation Details}
The US images are obtained by retrospectively masking the FS k-space \cite{fastmri}. We augment the datasets based on two types of under-sampling mask patterns - Cartesian and Gaussian masks with three different acceleration factors for under-sampling, namely, 4x, 5x, and 8x for training. We pose each configuration of contrast, mask type, and acceleration factor as a task dataset for the meta-learning models used for comparison. 
We have used the $L1$ loss function for optimization. 
The models are trained for 150 epochs. All models are implemented using PyTorch and trained on NvidiaRTX-3090 GPU with 24 GB memory. The models are implemented in PyTorch. The models have 5 cascades each. Adam optimizer is used with a learning rate of 0.001.

\subsection{Results}


\begin{figure*}[t!]
    \centering
    \includegraphics[width=\linewidth]{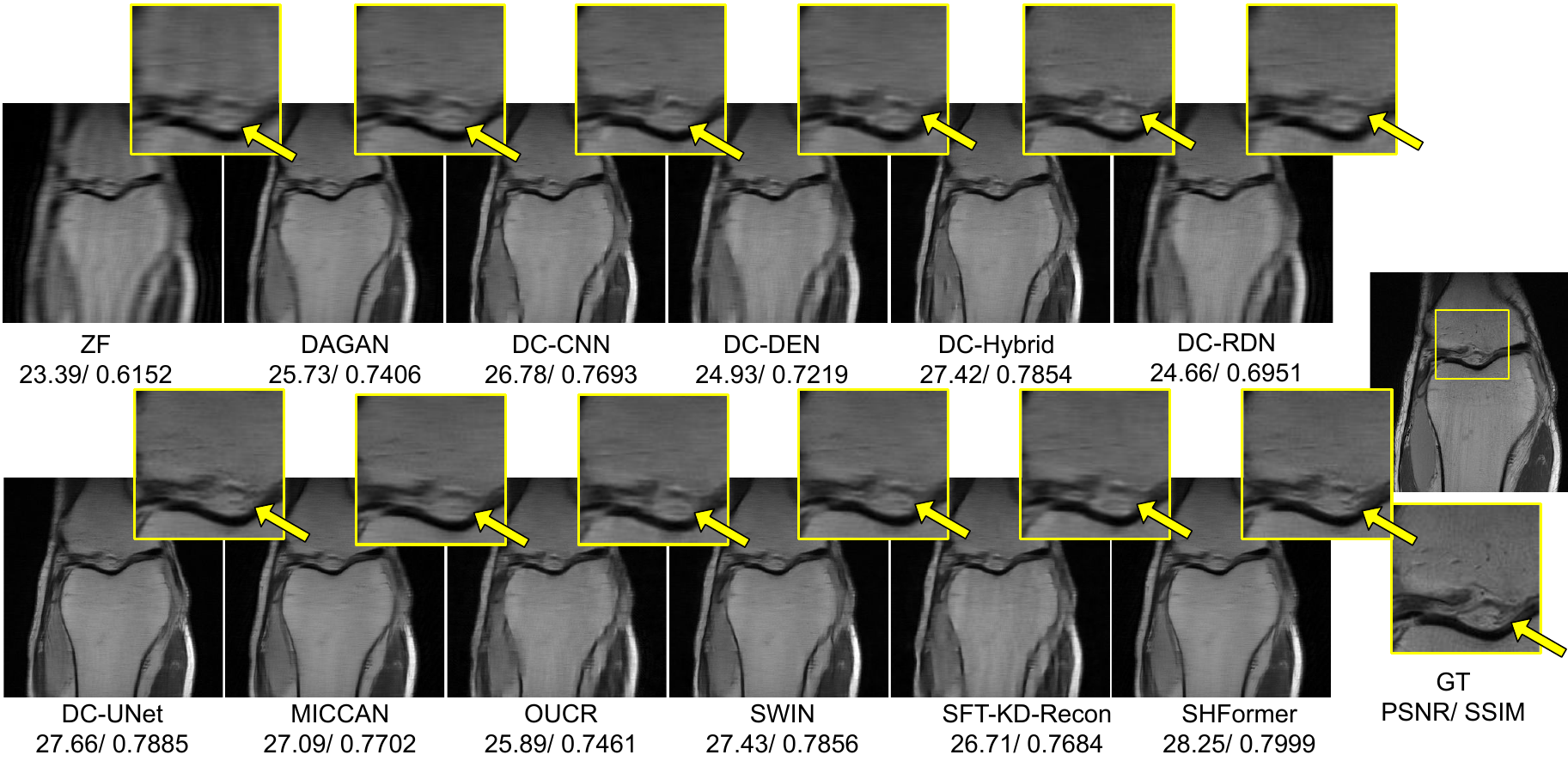}
    \caption{Qualitative comparison of the predictions of SHFormer with other reconstructed images. The figures show the predictions for 8x knee MRI reconstruction.  The quality of the proposed model is better than other reconstruction methods as highlighted near the ligament regions in the knee.
    }
    \label{fig:sc_knee}
\end{figure*}

\subsubsection{Evaluation on large-scale clinical datasets}

We compare our model against other neural network architectures proposed for single-coil image reconstruction using the ACDC cardiac challenge dataset and the realistic multi-scanner complex-valued fastMRI knee dataset. This experiment aims to highlight the superiority of our approach when the training data is abundant and where the baseline models can also perform competitively.  Table \ref{tab:sc_cardiac_knee} shows the quantitative comparison of SHFomer with other networks - CNN-based, DC-CNN \cite{dc_cnn}, DenseNet, DC-DEN \cite{dc-ensemble}, residual dilated network \cite{recursive_dilated}, U-Net \cite{dc_unet}, channel-attention-based, MICCAN \cite{miccan}, DC-hybrid (k-space and image-based) \cite{hybrid},  GAN-based, DAGAN \cite{dagan}, Swin transformer \cite{swin} and KD-based approach - SFT-KD-Recon \cite{sftkdrecon}. 

Figures \ref{fig:sc_cardiac} and \ref{fig:sc_knee} show the visual comparison of SHFormer against these methods with a focus on regions near the ventricles for cardiac and around the ligament regions of the knee, respectively. From the table and the visual results, our observations are as follows. (i) SHFormer gives high accuracy metrics over other methods. (ii) The improvements are consistent across acceleration factors - 4x, 5x, and 8x in cardiac scans, where the resolution of the images is limited, and for knee scans with higher resolution. (iii) The best improvement margin of SHFormer over the competitive model MICCAN is 0.5 dB PSNR and 0.006 SSIM. (iv) The visual results show that SHFormer exhibits superior reconstruction quality over other deep neural network approaches. The regions of interest in cardiac MRI reconstruction show better structural recovery, while the knee MRI shows better recovery of both structures and fine-grained features in the image.
We conjecture that the proposed spectral filtering and high-pass kernel generation modules of SHFormer provide a soft attention mechanism focusing on low-level and high-level visual features. Combining these features provides different levels of attention to varied concepts, enabling these features to contribute differently towards the overall reconstruction.

\begin{figure*}[t!]
    \centering
    \includegraphics[width=\linewidth]{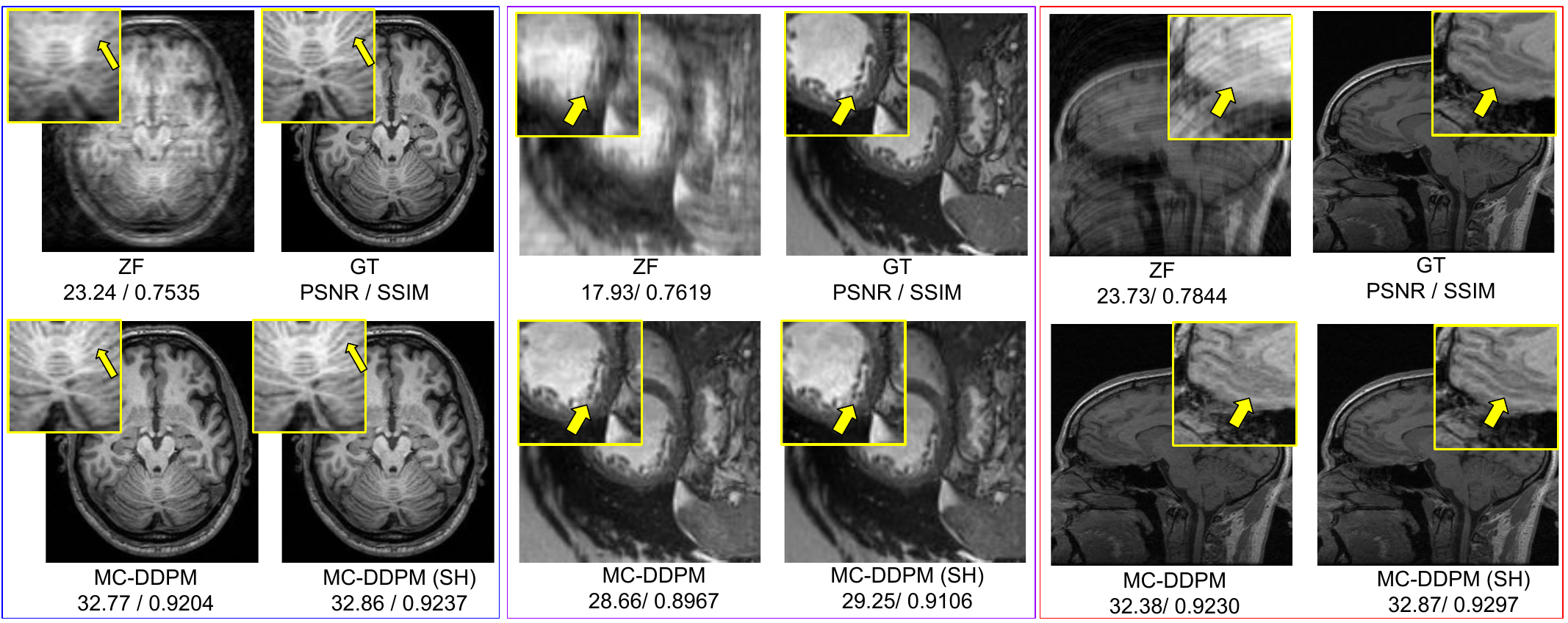}
    \caption{Qualitative comparison of the predictions of MC-DDPM with and without the SF and HF layers for 5x under-sampling in MRBrains - axial brain T1 MRI (left), 
    \noindent ACDC cardiac short axis view (middle), and Calgary brain image (right). In all three cases, the  
    highlighted patterns are much clearer in the proposed diffusion model setting compared to the baseline DDPM. 
    }
    \label{fig:mcddpm_rev}
\end{figure*}


\begin{table*}[t!]
\scriptsize
\centering
\caption{Quantitative comparison of MC-DDPM \cite{mcddpm} with and without the special filtering and high-pass kernel generation blocks. MC-DDPM (SH) denotes the diffusion model with spectral filtering modules and the high pass kernel generation transformer. The results are shown for three different datasets across multiple anatomies and acceleration factors.}
\label{tab:ddpm_nn_rev}
\setlength{\tabcolsep}{4pt}
\renewcommand{\arraystretch}{1.5}
\begin{tabular}{llcccc}
\hline
\multicolumn{1}{c}{\multirow{2}{*}{Dataset}}                             & \multicolumn{1}{c}{\multirow{2}{*}{Model}} & \multicolumn{2}{c}{4x}               & \multicolumn{2}{c}{5x}                 \\ \cline{3-6} 
\multicolumn{1}{c}{}                                                     & \multicolumn{1}{c}{}                       & PSNR             & SSIM              & PSNR              & SSIM               \\ \hline
\multirow{2}{*}{MRBrains}                                                & MC-DDPM                                    & 35.31 $\pm$ 5.50 & 0.9594 $\pm$ 0.03 & 34.68 $\pm$ 6.00  & 0.9480 $\pm$ 0.04  \\
                                                                         & MC-DDPM (SH)                              & 35.33 $\pm$ 5.72 & 0.9623 $\pm$ 0.03 & 34.86 $\pm$ 6.18  & 0.9518 $\pm$ 0.04  \\ \hline
\multirow{2}{*}{\begin{tabular}[c]{@{}l@{}}Calgary \\ brain\end{tabular}}                                                 & MC-DDPM                                    & 33.33 $\pm$ 1.71 & 0.9108 $\pm$ 0.02 & 31.91 $\pm$ 1.70  & 0.8458 $\pm$ 0.02  \\
                                                                         & MC-DDPM (SH)                              & 33.35 $\pm$ 1.69 & 0.9154 $\pm$ 0.02 & 31.97 $\pm$ 1.68  & 0.8485 $\pm$ 0.02  \\ \hline
\multirow{2}{*}{\begin{tabular}[c]{@{}l@{}}ACDC \\ cardiac\end{tabular}} & MC-DDPM                                    & 31.16 $\pm$ 3.67 & 0.8815 $\pm$ 0.06 & 30.17 $\pm$  3.75 & 0.8606 $\pm$  0.06 \\
                                                                         & MC-DDPM (SH)                              & 31.25 $\pm$ 3.72 & 0.8896 $\pm$ 0.05 & 30.21 $\pm$  3.82 & 0.8642 $\pm$  0.06 \\ \hline
\end{tabular}
\end{table*}

\begin{table}[t!]
\scriptsize
\centering
Computational cost analysis metrics of MC-DDPM (SH) relative to MC-DDPM in terms of the number of parameters, number of FLOPS and the mean sampling time during inference. Here, Difference and Relative increase denote the differences and the relative increase in these metrics in MC-DDPM (SH) with respect to MC-DDPM.
\label{tab:computecost}
\setlength{\tabcolsep}{2pt}
\renewcommand{\arraystretch}{1.5}
\begin{tabular}{lcccc}
\hline
\multicolumn{1}{c}{\multirow{2}{*}{Metric}} & \multirow{2}{*}{\# parameters} & \multirow{2}{*}{\begin{tabular}[c]{@{}c@{}}\# FLOPs\\ (GFLOPs)\end{tabular}} & \multicolumn{2}{c}{\begin{tabular}[c]{@{}c@{}}Mean Sampling time\\ (seconds)\end{tabular}} \\ \cline{4-5} 
\multicolumn{1}{c}{}                        &                                &                                                                              & Cardiac                                   & Brain                                     \\ \hline
MC-DDPM                                     & 3663778                        & 27.2685                                                                        & 12.32                                     & 12.25                                     \\
MC-DDPM (SH)                                & 3703055                        & 27.3061                                                                        & 13.72                                     & 14.16                                     \\
Difference                                  & 39277                          & 0.05                                                                         & 1.4                                       & 1.91                                      \\
Relative increase                           & 1.07\%                        & 0.138\%                                                                      & 11.36\%                                   & 15.51\%                                   \\ \hline
\end{tabular}
\end{table}


\begin{table}[t!]
\centering
\scriptsize
\caption{Quantitative comparison of SHFormer with other multi-coil MRI reconstruction methods using axial T2 knee MRI dataset for physics-driven self-supervised learning }
\label{tab:mc_axialt2}
\setlength{\tabcolsep}{3pt}
\renewcommand{\arraystretch}{1.15}
\begin{tabular}{@{}ccccc@{}}
\toprule
\multirow{2}{*}{\textbf{Method}} & \multicolumn{2}{c}{\textbf{4x}}                        & \multicolumn{2}{c}{\textbf{5x}}                        \\ \cmidrule(l){2-3} \cmidrule(l){4-5} 
                                 & \textbf{PSNR}             & \textbf{SSIM}              & \textbf{PSNR}             & \textbf{SSIM}              \\ \midrule
\textbf{ZF}                      & 31.35 $\pm$ 3.69          & 0.8186 $\pm$ 0.06          & 30.38 $\pm$ 3.76          & 0.7829 $\pm$ 0.07          \\
\textbf{VSNet}                   & 32.17 $\pm$ 3.44          & 0.8269 $\pm$ 0.06          & 30.82 $\pm$ 3.60          & 0.7885 $\pm$ 0.07          \\
\textbf{PDN }                     & 33.73 $\pm$ 3.26          & 0.8501 $\pm$ 0.05          & 31.59 $\pm$ 3.44          & 0.7988 $\pm$ 0.07          \\
\textbf{RecVarnet}               & 32.44 $\pm$ 3.48          & 0.8402 $\pm$ 0.05          & 31.26 $\pm$ 3.28          & 0.7983 $\pm$ 0.06          \\
\textbf{KIKI-Net}                & 34.12 $\pm$ 3.17          & 0.8564 $\pm$ 0.05          & 32.65 $\pm$ 3.04          & 0.8138 $\pm$ 0.06          \\
\textbf{UNet}                    & 34.36 $\pm$ 3.07          & 0.8596 $\pm$ 0.05          & 32.67 $\pm$ 3.07          & 0.8152 $\pm$ 0.06          \\
\textbf{ISTA}                    & 33.73 $\pm$ 3.26          & 0.8485 $\pm$ 0.05          & 31.82 $\pm$ 3.31          & 0.8013 $\pm$ 0.07          \\
\textbf{SWIN}                    & 34.38 $\pm$ 3.06          & 0.8596 $\pm$ 0.05          & 32.81 $\pm$ 3.10          & 0.8157 $\pm$ 0.06          \\
\textbf{SHFormer}                & \textbf{34.71 $\pm$ 3.16} & \textbf{0.8691 $\pm$ 0.04} & \textbf{32.85 $\pm$ 3.35} & \textbf{0.8263 $\pm$ 0.06} \\ \bottomrule
\end{tabular}
\end{table}

\subsubsection{Diffusion model-based study}

We compare the proposed MC-DDPM (SH) with the baseline MC-DDPM diffusion models to assess the contribution of dynamic spectral and high-pass filtering in the generative model setting.
The quantitative and visual comparison of MC-DDPM (SH) with MC-DDPM are shown in Table \ref{tab:ddpm_nn_rev} and Figure \ref{fig:mcddpm_rev}, respectively for MRBrains and Calgary T1 brain datasets, and ACDC cardiac datasets with 4x and 5x Cartesian acceleration factors. From the table, we note that (i) the proposed diffusion model consistently performs better than the baseline DDPM model across different anatomies, acceleration factors, and dataset types, and (ii) the best improvement of 0.14 dB in PSNR is obtained for MRBrain 5x and 0.0081 in SSIM for cardiac 4x.
The consistent improvement in performance across acceleration factors reveals the significance of our architectural choices with dynamic spectral filtering and high-pass kernel generation in the attention blocks of the denoising diffusion model.  The proposed diffusion model incorporates scalability through self-conditioning using attention maps and modulating the maps without requiring external information during the reverse process. This approach can encouragingly boost the quality of the reconstruction with strong performance across all undersampling factors over the baseline.

\textbf{Computation cost analysis: }
Table \ref{tab:computecost} shows three perspectives of computation cost - the number of network parameters, the number of floating point operations (FLOPS), and the mean sampling time of the model at inference. From the table, our observations are as follows. (i) MC-DDPM (SH) shows a subtle increase in the number of parameters, FLOPs, and the mean sampling time. (ii) The relative increase in the number of parameters is 1.07\%, and that in the number of FLOPS is 0.14\%. Table \ref{tab:ddpm_nn_rev} shows that the best average improvement is obtained for the cardiac dataset with a margin of 0.008 ($\sim$ 0.01), and the second best margin of $\sim$ 0.004 for brain datasets. This improvement gives a best relative increase in SSIM of nearly 1\% and an average improvement of 0.5\%. This implies that, for the cardiac case, the error has reduced by $\sim$ 6.72\% in SSIM for an average increase of 0.14\% in the FLOPS and 13.4\% in the mean sampling time.


\begin{figure*}[t!]
    \centering
    \includegraphics[width=\linewidth]{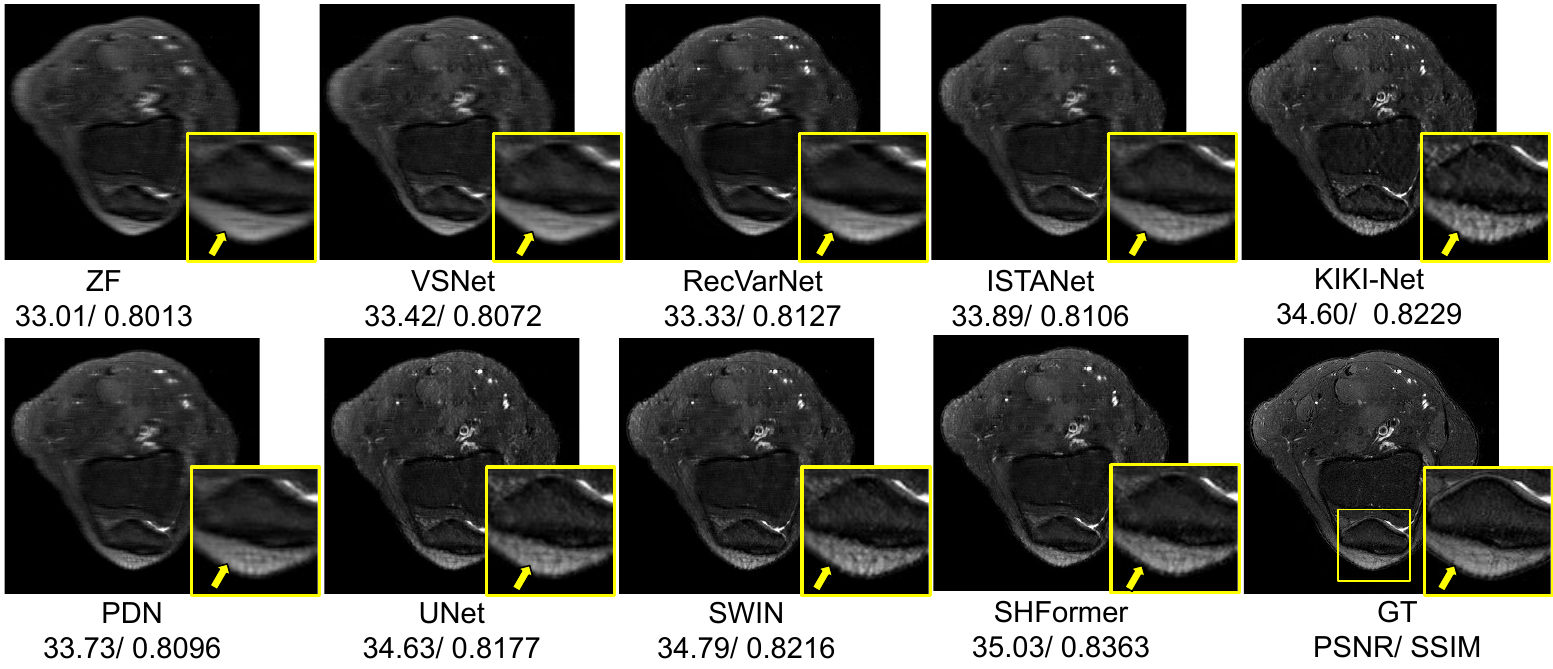}
    \caption{Qualitative comparison of the predictions of SHFormer with those of other methods for physics-driven self-supervised 4x axial T2 knee multi-coil MRI reconstruction.  The proposed method recovers textural details much closer to the target than the competitive baselines KIKINet and Swin transformer which exhibit noisy structures near the patella in the reconstruction.
    }
    \label{fig:mc_kneeaxial_t2}
\end{figure*}

\begin{table}[t!]
\centering
\scriptsize
\caption{Quantitative comparison of SHFormer with other multi-coil MRI reconstruction methods using coronal PD MRI data for physics-driven self-supervised learning}
\label{tab:mc_coronalpd}
\setlength{\tabcolsep}{3pt}
\renewcommand{\arraystretch}{1.15}
\begin{tabular}{@{}ccccc@{}}
\toprule
\multirow{2}{*}{\textbf{Method}} & \multicolumn{2}{c}{\textbf{4x}}                        & \multicolumn{2}{c}{\textbf{5x}}                        \\ \cmidrule(l){2-3} \cmidrule(l){4-5}
                                 & \textbf{PSNR}             & \textbf{SSIM}              & \textbf{PSNR}             & \textbf{SSIM}              \\ \midrule
\textbf{ZF}                      & 28.14 $\pm$ 3.72          & 0.7838 $\pm$ 0.09          & 25.99 $\pm$ 3.72          & 0.7119 $\pm$ 0.12          \\
\textbf{VSNet}                   & 30.65 $\pm$ 4.14          & 0.8431 $\pm$ 0.08          & 26.71 $\pm$ 3.60          & 0.7369 $\pm$ 0.10          \\
\textbf{PDN}                     & 32.10 $\pm$ 3.02          & 0.8698 $\pm$ 0.05          & 27.66 $\pm$ 3.43          & 0.7620 $\pm$ 0.10          \\
\textbf{RecVarnet}               & 29.49 $\pm$ 3.07          & 0.8255 $\pm$ 0.06          & 25.65 $\pm$ 2.92          & 0.7134 $\pm$ 0.10          \\
\textbf{KIKI-net}                & 31.80 $\pm$ 4.05          & 0.8617 $\pm$ 0.07          & 27.42 $\pm$ 3.63          & 0.7574 $\pm$ 0.10          \\
\textbf{UNet}                    & 32.90 $\pm$ 2.72          & 0.8884 $\pm$ 0.04          & 27.90 $\pm$ 3.73          & 0.7729 $\pm$ 0.09          \\
\textbf{ISTA-net}                & 31.94 $\pm$ 3.11          & 0.8635 $\pm$ 0.06          & 27.72 $\pm$ 3.48          & 0.7649 $\pm$ 0.10          \\
\textbf{SWIN}                    & 33.22 $\pm$ 2.88          & 0.8954 $\pm$ 0.05          & 28.43 $\pm$ 3.46          & 0.7853 $\pm$ 0.09          \\
\textbf{SHFormer}                & \textbf{33.57 $\pm$ 2.76} & \textbf{0.9009 $\pm$ 0.04} & \textbf{28.82 $\pm$ 3.49} & \textbf{0.7974 $\pm$ 0.08} \\ \bottomrule
\end{tabular}
\end{table}

\begin{figure*}[t!]
    \centering
    \includegraphics[width=\linewidth]{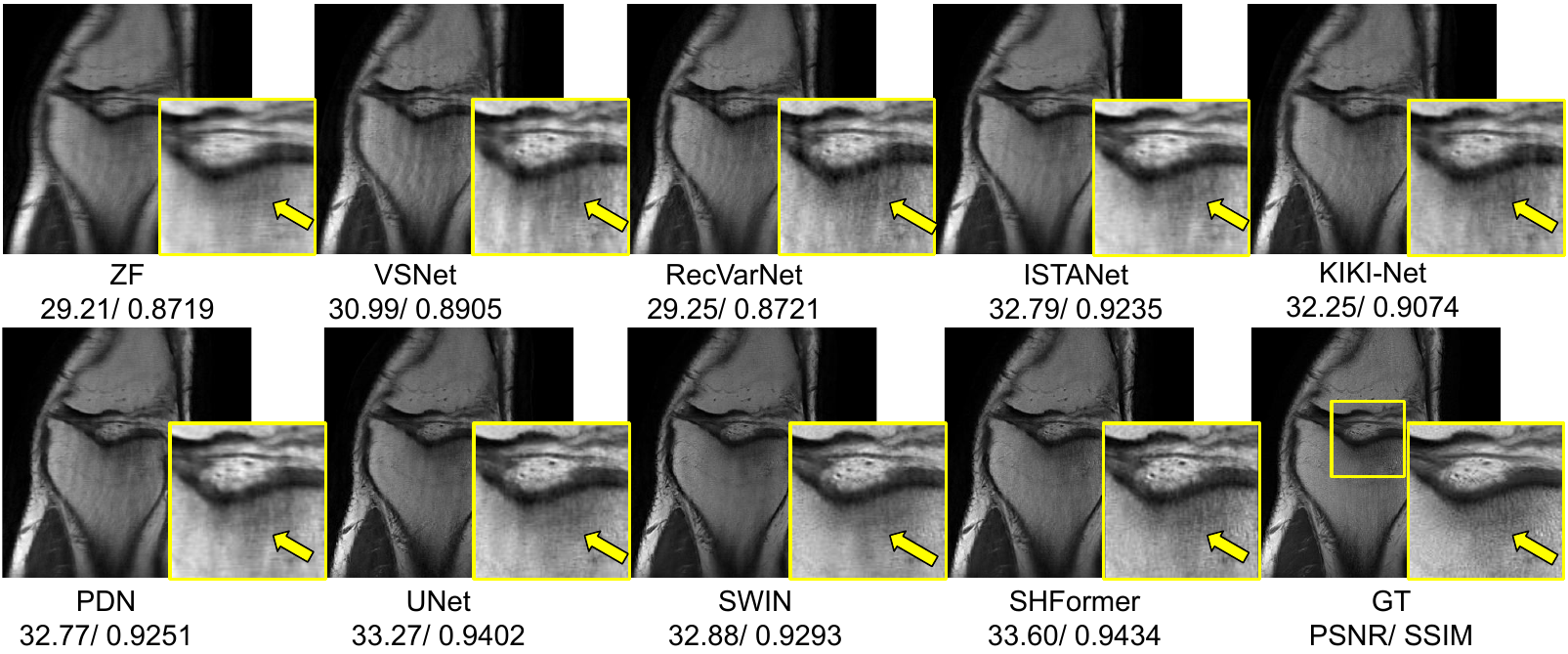}
    \caption{Qualitative comparison of the predictions of SHFormer with those of other methods for physics-driven self-supervised 4x coronal PD knee MRI multi-coil MRI reconstruction. The highlighted region shows the superior recovery of textural details on the bone regions for the proposed method over other methods.
    }
    \label{fig:mc_kneepd}
\end{figure*}

\subsubsection{Self-Supervised Learning for Multi-coil MRI Reconstruction}
Due to the physiological limitations and longer scan times of MRI acquisition for high-resolution fully-sampled measurements, self-supervised learning methods have gained attention to learn from under-sampled measurements without imposing strong
priors and without knowing the ground-truth signals \cite{yaman2020self}. Hence, we evaluate the performance of our method on physics-driven self-supervised image reconstruction for multi-coil MRI. We compare our method against other methods in multi-coil reconstruction namely variable splitting network (VSNet) \cite{vs_net}, recurrent variational network (RecVarnet) \cite{RecVarnet}, self-supervised parallel domain network (PDN) \cite{pdn}, hybrid k-space and image space network (KIKINet) \cite{kiki}, UNet \cite{unet}, interpretable optimization-inspired compressive sensing network (ISTA) \cite{ista} and Swin transformer \cite{swin}.
The quantitative evaluation against other multi-coil MRI reconstruction architectures for two contrasts, namely axial T2 and coronal PD, respectively, in Tables \ref{tab:mc_axialt2} and  \ref{tab:mc_coronalpd} show that SHFormer outperforms other methods in PSNR
and SSIM metrics on the MRI sequences. The average improvement margins in SHFormer over the immediate baseline methods, the SWIN transformer and the UNet, are 0.2 to 0.4 dB PSNR and around 0.004 in SSIM. The best improvement margins of around 0.6 dB PSNR and 0.01 in SSIM are obtained in the case of coronal PD. The visual results in Figures \ref{fig:mc_kneeaxial_t2} and \ref{fig:mc_kneepd} show better recovery of fine details,
 at the same time, previous methods, such as UNet and SWIN transformer, exhibit a noticeable amount of noise (Figure \ref{fig:mc_kneeaxial_t2}), and aliasing artifacts (Figure \ref{fig:mc_kneepd}). 

 These observations show that our implicit multi-level attention mechanism provided within the CNN and the transformers work collaboratively spatially and channel-wise to capture useful information and high-frequency details, extract long-range dependencies, and refine the output for better representation.

\begin{figure*}[t!]
    \centering
    \includegraphics[width=0.8\linewidth,height=0.3\linewidth]{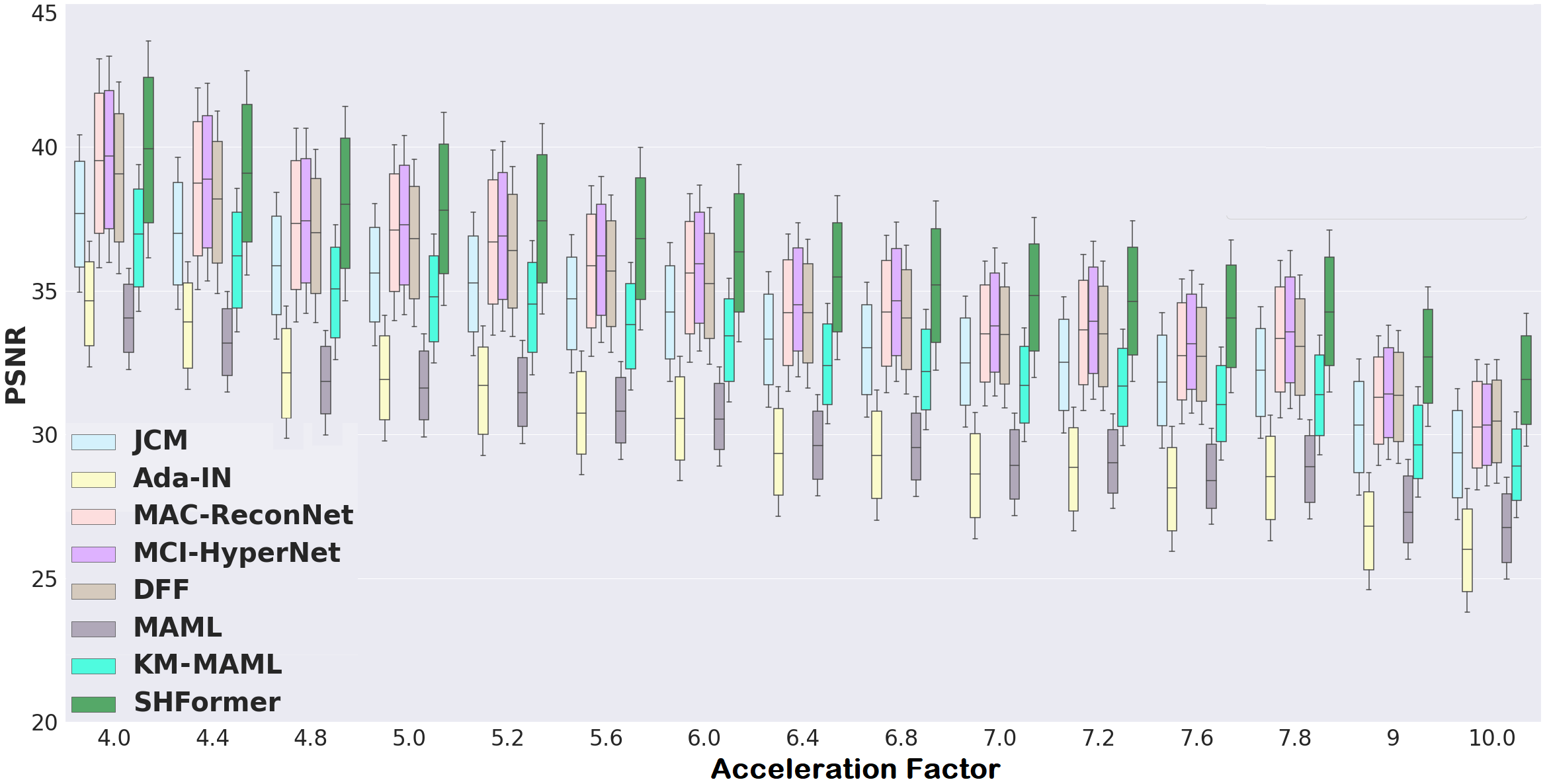}
    \caption{Plot comparing the PSNR metrics for the closed-set generalization capabilities of SHFormer with that of joint training, other adaptive learning networks, and meta-learning approaches for several unseen acceleration factors 
 from 4x to 10x when trained on a few acceleration factors. The plots show consistent and superior performance over adaptive learning approaches, showcasing scalability to varying unseen under-sampling levels.
    }
    \label{fig:boxpsnr}
\end{figure*}

\begin{figure*}[t!]
    \centering
    \includegraphics[width=0.8\linewidth]{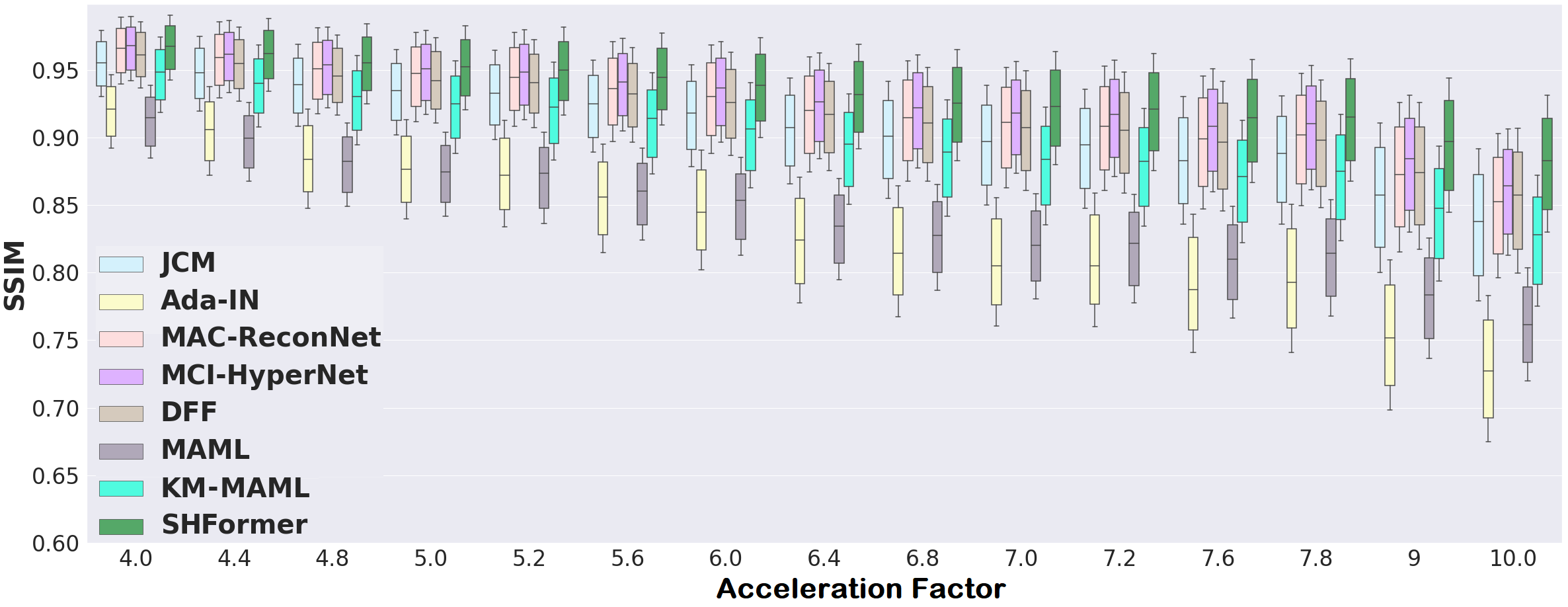}
    \caption{Plot comparing the SSIM metrics for the closed-set generalization capabilities of SHFormer with that of joint training, other adaptive learning networks, and meta-learning approaches for several unseen acceleration factors from 4x to 10x when trained on a few acceleration factors. The SSIM plot indicates that the proposed method has the potential to minimize retraining owing to its ability to tune to continuously varying acceleration factors.
    }
    \label{fig:boxssim}
\end{figure*}

\begin{table*}[t!]
\centering
\scriptsize
\caption{Quantitative Comparison of the closed-set generalization performance of SHFormer with joint learning or joint context model (JCM), adaptive models - adaptive instance normalization (Ada-IN), decoupled learning (MAC-ReconNet), coupled learning (MCI-HyperNet), deep frequency filtering (DFF), vanilla model-agnostic meta-learning (MAML), kernel modulation-based model-agnostic meta-learning (KM-MAML). AF denotes acceleration factor.}
\label{tab:closedset}
\setlength{\tabcolsep}{7pt}
\renewcommand{\arraystretch}{1.25}
\begin{tabular}{ccccccccc}
\hline
\multirow{2}{*}{AF} & \multicolumn{2}{c}{JCM}              & \multicolumn{2}{c}{MAC-ReconNet \cite{mac}}     & \multicolumn{2}{c}{MCI-HyperNet \cite{mcihypernet}}     & \multicolumn{2}{c}{DFF \cite{dff}}                                \\ \cmidrule(lr){2-3} \cmidrule(lr){4-5} \cmidrule(lr){6-7} \cmidrule(lr){8-9}
                    & PSNR             & SSIM              & PSNR             & SSIM              & PSNR             & SSIM              & PSNR                      & SSIM                       \\ \hline
3x                  & 40.20 $\pm$ 4.69 & 0.9796 $\pm$ 0.01 & 42.02 $\pm$ 5.83 & 0.9847 $\pm$ 0.01 & 42.12 $\pm$ 5.78 & 0.9853 $\pm$ 0.01 & 41.37 $\pm$ 5.11          & 0.9828 $\pm$ 0.01          \\
4.2x                & 36.98 $\pm$ 3.92 & 0.9610 $\pm$ 0.02 & 38.65 $\pm$ 5.06 & 0.9698 $\pm$ 0.02 & 38.82 $\pm$ 5.04 & 0.9717 $\pm$ 0.02 & 38.16 $\pm$ 4.45          & 0.9672 $\pm$ 0.02          \\
5x                  & 35.33 $\pm$ 3.65 & 0.9472 $\pm$ 0.03 & 36.74 $\pm$ 4.74 & 0.9577 $\pm$ 0.03 & 36.98 $\pm$ 4.70 & 0.9610 $\pm$ 0.03 & 36.43 $\pm$ 4.17          & 0.9550 $\pm$ 0.03          \\
5.6x                & 34.34 $\pm$ 3.62 & 0.9361 $\pm$ 0.03 & 35.48 $\pm$ 4.51 & 0.9476 $\pm$ 0.03 & 35.80 $\pm$ 4.54 & 0.9522 $\pm$ 0.03 & 35.36 $\pm$ 4.01          & 0.9452 $\pm$ 0.03          \\
6.2x                & 33.47 $\pm$ 3.61 & 0.9239 $\pm$ 0.04 & 34.65 $\pm$ 4.27 & 0.9371 $\pm$ 0.04 & 34.93 $\pm$ 4.27 & 0.9424 $\pm$ 0.04 & 34.40 $\pm$ 3.94          & 0.9341 $\pm$ 0.04          \\
7x                  & 32.18 $\pm$ 3.63 & 0.9076 $\pm$ 0.04 & 33.25 $\pm$ 4.06 & 0.9228 $\pm$ 0.05 & 33.54 $\pm$ 4.01 & 0.9298 $\pm$ 0.04 & 33.19 $\pm$ 3.82          & 0.9202 $\pm$ 0.04          \\
7.8x                & 31.89 $\pm$ 3.58 & 0.8992 $\pm$ 0.05 & 33.00 $\pm$ 4.07 & 0.9150 $\pm$ 0.05 & 33.33 $\pm$ 4.13 & 0.9238 $\pm$ 0.04 & 32.78 $\pm$ 3.76          & 0.9119 $\pm$ 0.05          \\ \hline
\multirow{2}{*}{AF} & \multicolumn{2}{c}{Ada-IN \cite{univusmri}}           & \multicolumn{2}{c}{MAML \cite{maml}}             & \multicolumn{2}{c}{KM-MAML \cite{kmmaml}}          & \multicolumn{2}{c}{SHFormer}                           \\ \cmidrule(lr){2-3} \cmidrule(lr){4-5} \cmidrule(lr){6-7} \cmidrule(lr){8-9} 
                    & PSNR             & SSIM              & PSNR             & SSIM              & PSNR             & SSIM              & PSNR                      & SSIM                       \\ \hline
3x                  & 37.85 $\pm$ 3.62 & 0.9669 $\pm$ 0.01 & 37.23 $\pm$ 3.57 & 0.9608 $\pm$ 0.02 & 39.24 $\pm$ 4.52 & 0.9757 $\pm$ 0.01 & \textbf{42.23 $\pm$ 5.82} & \textbf{0.9853 $\pm$ 0.01} \\
4.2x                & 33.74 $\pm$ 3.38 & 0.9238 $\pm$ 0.03 & 33.30 $\pm$ 3.13 & 0.9179 $\pm$ 0.03 & 36.18 $\pm$ 3.97 & 0.9547 $\pm$ 0.03 & \textbf{39.11 $\pm$ 5.13} & \textbf{0.9725 $\pm$ 0.02} \\
5x                  & 31.67 $\pm$ 3.49 & 0.8933 $\pm$ 0.04 & 31.48 $\pm$ 3.07 & 0.8909 $\pm$ 0.03 & 34.47 $\pm$ 3.71 & 0.9386 $\pm$ 0.03 & \textbf{37.54 $\pm$ 4.80} & \textbf{0.9633 $\pm$ 0.03} \\
5.6x                & 30.49 $\pm$ 3.51 & 0.8706 $\pm$ 0.05 & 30.64 $\pm$ 3.10 & 0.8739 $\pm$ 0.04 & 33.49 $\pm$ 3.62 & 0.9267 $\pm$ 0.04 & \textbf{36.52 $\pm$ 4.59} & \textbf{0.9556 $\pm$ 0.03} \\
6.2x                & 29.66 $\pm$ 3.37 & 0.8469 $\pm$ 0.05 & 29.84 $\pm$ 3.01 & 0.8539 $\pm$ 0.04 & 32.57 $\pm$ 3.51 & 0.9130 $\pm$ 0.04 & \textbf{35.53 $\pm$ 4.38} & \textbf{0.9465 $\pm$ 0.03} \\
7x                  & 28.31 $\pm$ 3.44 & 0.8176 $\pm$ 0.06 & 28.79 $\pm$ 3.07 & 0.8325 $\pm$ 0.05 & 31.47 $\pm$ 3.49 & 0.8972 $\pm$ 0.05 & \textbf{34.50 $\pm$ 4.17} & \textbf{0.9366 $\pm$ 0.04} \\
7.8x                & 28.22 $\pm$ 3.38 & 0.8081 $\pm$ 0.06 & 28.65 $\pm$ 3.06 & 0.8253 $\pm$ 0.05 & 31.13 $\pm$ 3.45 & 0.8883 $\pm$ 0.05 & \textbf{33.99 $\pm$ 4.12} & \textbf{0.9295 $\pm$ 0.04} \\ \hline
\end{tabular}
\end{table*}

\begin{figure*}[t!]
    \centering
    \includegraphics[width=\linewidth]{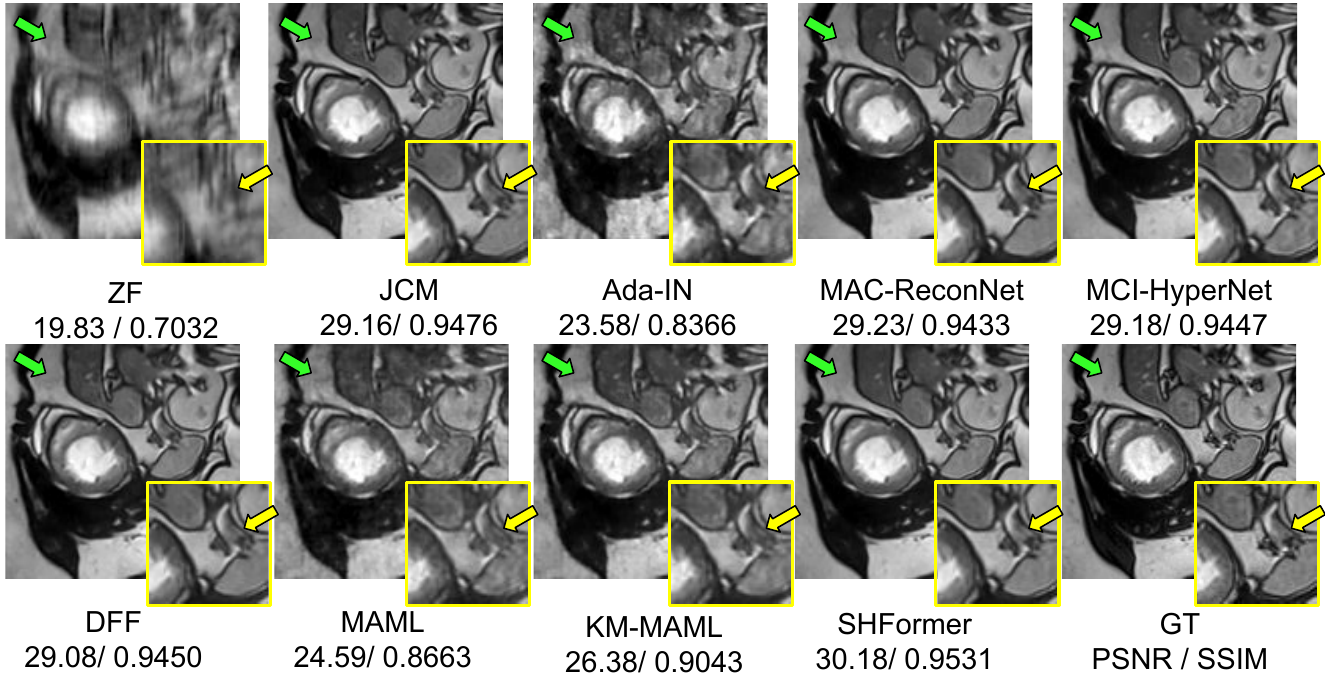}
    \caption{Qualitative comparison of the closed-set generalization capabilities of SHFormer with joint training, task-level adaptive approaches at the architecture, and model-agnostic meta-learning methods for cardiac 8x undersampling. The results show higher accuracy for SHFormer in recovering regions with subtle intensity changes (yellow arrows) and homogeneous regions (green arrows)
    }
    \label{fig:closedset}
\end{figure*}

\subsubsection{Closed-set Generalization}
In the closed-set DG experiment, we evaluate the proposed model with other adaptive methods (joint learning or joint context model \cite{mac} that combines images from different acceleration factors and trains a backbone network, task-level adaptive learning methods - MAC-ReconNet \cite{mac} and MCI-HyperNet \cite{mcihypernet}, instance-level AM methods - adaptive instance normalization \cite{univusmri} and DFF \cite{dff}, and optimization-based meta-learning methods -  MAML \cite{maml}, KM-MAML \cite{kmmaml} ) on domains where the label space is shared \cite{dff} across varying levels of under-sampling. The closed-set generalization can be perceived in the context of training on few acceleration factors (2x, 2.5x, 3.3x, 4x, 5x, and 8x) and evaluating the model on a wide range of  unseen acceleration factors from 2x to 10x in increments of 0.2 with unseen mask patterns (35 unseen acceleration factors).

Figures \ref{fig:boxpsnr} and \ref{fig:boxssim} show the PSNR and SSIM box plots that compare SHFormer with other adaptive methods for 15 unseen cases.
From the PSNR plots, we note that as high-frequency components are removed from the k-space along the acceleration factor scale from 6.0x to 9.0x, we see that competitive adaptive and meta-learning baselines, MCI-HyperNet and KM-MAML exhibit steeper drops in performance (steeper slope) from (i) 6.0x to 6.4x, (ii) 6.8x to 7.0x, and (iii) 7.2x to 7.6x when compared to SHFormer indicating their superiority in capturing low-frequency information. 
In the case of SSIM, SHFormer exhibits a limited drop in performance from 6.4x to 7.2x compared to other models. For example, SHFormer shows a drop of 0.003 in SSIM from 3.6x to 4.6x (from 0.9796 to 0.976), while MCI-HyperNet shows a higher drop of 0.014 (from 0.9794 to 0.9655). Furthermore, the improvements in SHFormer is much higher at higher acceleration factors over the competitive baselines DFF and MCI-HyperNet. These observations indicate SHFormer's superior ability to reconstruct high-frequency information from the low-frequency components and fewer high-frequency k-space measurements compared to the competitive adaptive baselines MAC-ReconNet, MCI-HyperNet, and DFF. Furthermore, the plots signify the method's potential in dynamically adjusting the filter weights, which provides a comprehensive tunable representation of unseen under-sampling factors in a single model without retraining.

Table \ref{tab:closedset} shows the quantitative metric comparison of SHFormer with other methods for seven unseen acceleration factors (AF). Figure \ref{fig:closedset} shows the visual comparison for 8x unseen cases.
The improved metrics and better visual results of SHFormer indicate that the proposed SFCNN and HKTFB leverage the higher receptive field of the sub-sampling layers in the CNN and the self-attention blocks of the transformer in ensembling reusable and high-frequency details. 



\begin{table*}[t!]
\centering
\scriptsize
\caption{Open-set generalization performance for multimodal (multi-contrast) MRI between SHFormer and other adaptive learning approaches. The first column indicates the MRI contrasts used during training and the unseen contrast evaluated at inference time. AF denotes acceleration factor. Mask types - 'C' denotes Cartesian under-sampling pattern, and 'G' denotes Gaussian undersampling patterns. }
\label{tab:openset}
\setlength{\tabcolsep}{4pt}
\renewcommand{\arraystretch}{1.25}
\begin{tabular}{ccccccccc}
\hline
\multirow{2}{*}{\begin{tabular}[c]{@{}c@{}}Seen $\rightarrow$\\  unseen domains\end{tabular}} & \multirow{2}{*}{\begin{tabular}[c]{@{}c@{}}Mask type\\ +\ AF\end{tabular}} & MAC-ReconNet   & MCI-HyperNet   & Ada-IN         & DFF            & MAML           & KM-MAML        & SHFormer       \\ \cline{3-9} 
                                                                                                           &                                                                             & PSNR / SSIM    & PSNR / SSIM    & PSNR / SSIM    & PSNR / SSIM    & PSNR / SSIM    & PSNR / SSIM    & PSNR / SSIM    \\ \hline
\multirow{4}{*}{T1, FL, PD $\rightarrow$ T2}                                                                  & C, 4x                                                                       & 34.05 / 0.9354 & 35.13 / 0.9565 & 35.34 / 0.9507 & 35.68 / 0.9586 & 31.27 / 0.8412 & 32.82 / 0.9186 & 36.01 / 0.9610 \\
                                                                                                           & C, 5x                                                                       & 33.39 / 0.9135 & 34.39 / 0.9413 & 34.69 / 0.9352 & 35.37 / 0.9474 & 29.52 / 0.7825 & 31.65 / 0.8865 & 35.56 / 0.9501 \\
                                                                                                           & G, 5x                                                                       & 39.62 / 0.9602 & 41.81 / 0.9786 & 41.95 / 0.9769 & 41.34 / 0.9745 & 33.70 / 0.8146 & 35.97 / 0.9097 & 42.50 / 0.9795 \\
                                                                                                           & G, 8x                                                                       & 34.98 / 0.9160 & 36.56 / 0.9436 & 36.80 / 0.9376 & 36.71 / 0.9411 & 30.12 / 0.7263 & 31.80 / 0.8374 & 37.93 / 0.9535 \\ \hline
\multirow{4}{*}{T1, T2, PD $\rightarrow$ FL}                                                                  & C, 5x                                                                       & 36.70 / 0.9581 & 37.64 / 0.9623 & 37.53 / 0.9635 & 38.50 /- 0.9713 & 32.44 / 0.8400 & 35.37 / 0.9346 & 38.56 / 0.9721 \\
                                                                                                           & G, 4x                                                                       & 47.16 / 0.9944 & 45.78 / 0.9885 & 48.71 / 0.9960 & 47.45 / 0.9929 & 38.84 / 0.9035 & 44.28 / 0.9786 & 48.74 / 0.9955 \\
                                                                                                           & G, 5x                                                                       & 45.00 / 0.9902 & 44.53 / 0.9849 & 46.35 / 0.9924 & 45.30 / 0.9884 & 36.35 / 0.8666 & 41.02 / 0.9572 & 46.49 / 0.9917 \\
                                                                                                           & G, 8x                                                                       & 39.86 / 0.9712 & 40.57 / 0.9691 & 41.05 / 0.9768 & 40.37 / 0.9714 & 32.99 / 0.8027 & 36.62 / 0.9240 & 41.58 / 0.9786 \\ \hline
\multirow{4}{*}{T1, FL, T2 $\rightarrow$ PD}                                                                  & C, 4x                                                                       & 35.33 / 0.9424 & 36.87 / 0.9639 & 35.49 / 0.9485 & 36.71 / 0.9622 & 31.75 / 0.8508 & 34.88 / 0.9453 & 36.99 / 0.9643 \\
                                                                                                           & C, 5x                                                                       & 34.66 / 0.9235 & 35.89 / 0.9489 & 34.56 / 0.9279 & 35.98 / 0.9478 & 29.93 / 0.7948 & 33.63 / 0.9235 & 36.17 / 0.9522 \\
                                                                                                           & C, 8x                                                                       & 30.25 / 0.8755 & 31.85 / 0.9087 & 30.15 / 0.8760 & 32.21 / 0.9158 & 27.80 / 0.7596 & 30.23 / 0.8821 & 32.23 / 0.9160 \\
                                                                                                           & G, 4x                                                                       & 43.70 / 0.9803 & 46.07 / 0.9889 & 46.13 / 0.9887 & 43.96 / 0.9855 & 36.70 / 0.8861 & 42.33 / 0.9767 & 46.13 / 0.9898 \\ \hline
\end{tabular}
\end{table*}

\begin{table*}[t!]
\centering
\scriptsize
\caption{Open-set Generalization performance for multimodal MRI between SHFormer and other adaptive learning approaches for \textbf{\underline{complex-valued}} dataset. "Seen" domain - the training and test data domains match (coronal PD and PDFS, 4x acceleration). "Unseen" and "Unseen (Adapt)" - adaptation on the fly and via fine-tuning to coronal PDFS, respectively when trained on coronal PD images. SHFormer (hyp) - only the spectral and high-pass attention hypernetworks are finetuned.}
\label{tab:multimodal_complex_adapt}
\setlength{\tabcolsep}{4pt}
\renewcommand{\arraystretch}{1.25}
\begin{tabular}{lcccccc}
\hline
\multicolumn{1}{c}{\multirow{3}{*}{Method}} & \multicolumn{6}{c}{Data domain}                                                                                    \\ \cline{2-7} 
\multicolumn{1}{c}{}                        & \multicolumn{2}{c}{Seen}             & \multicolumn{2}{c}{Unseen}           & \multicolumn{2}{c}{Unseen (Adapt)}   \\ \cline{2-7} 
\multicolumn{1}{c}{}                        & PSNR             & SSIM              & PSNR             & SSIM              & PSNR             & SSIM              \\ \hline
MAC-ReconNet                                & 29.62 $\pm$ 4.22 & 0.7958 $\pm$ 0.16 & 28.45 $\pm$ 5.49 & 0.6964 $\pm$ 0.18 & 28.75 $\pm$ 5.88 & 0.7066 $\pm$ 0.19 \\
MAML                                        & 30.21 $\pm$ 3.93 & 0.8350 $\pm$ 0.13 & 29.70 $\pm$ 4.73 & 0.7689 $\pm$ 0.15 & 30.00 $\pm$ 5.02 & 0.7769 $\pm$ 0.15 \\
Ada-IN                                      & 30.78 $\pm$ 4.33 & 0.8421 $\pm$ 0.15 & 29.71 $\pm$ 5.11 & 0.7593 $\pm$ 0.16 & 29.96 $\pm$ 5.52 & 0.7672 $\pm$ 0.17 \\
DFF                                         & 29.68 $\pm$ 4.45 & 0.8000 $\pm$ 0.17 & 28.51 $\pm$ 5.46 & 0.6981 $\pm$ 0.18 & 28.55 $\pm$ 5.82 & 0.7016 $\pm$ 0.19 \\
KM-MAML                                    & 31.07 $\pm$ 4.80 & 0.8513 $\pm$ 0.15 & 30.05 $\pm$ 5.14 & 0.7802 $\pm$ 0.15 & 30.47 $\pm$ 5.44 & 0.7910 $\pm$ 0.16 \\
MCI-HyperNet                               & 31.06 $\pm$ 4.57 & 0.8521 $\pm$ 0.14 & 30.14 $\pm$ 5.26 & 0.7829 $\pm$ 0.16 & 30.61 $\pm$ 5.52 & 0.7924 $\pm$ 0.16 \\
SHFormer                                    & 31.24 $\pm$ 4.55 & 0.8568 $\pm$ 0.15 & 30.67 $\pm$ 5.28 & 0.7891 $\pm$ 0.16 & 30.76 $\pm$ 5.59 & 0.7988 $\pm$ 0.16 \\
SHFormer (hyp)                    &                  &                   &                  &                   & 30.70 $\pm$ 5.51 & 0.7951 $\pm$ 0.16 \\ \hline
\end{tabular}
\end{table*}

\begin{figure*}[t!]
    \centering
    \includegraphics[width=\linewidth]{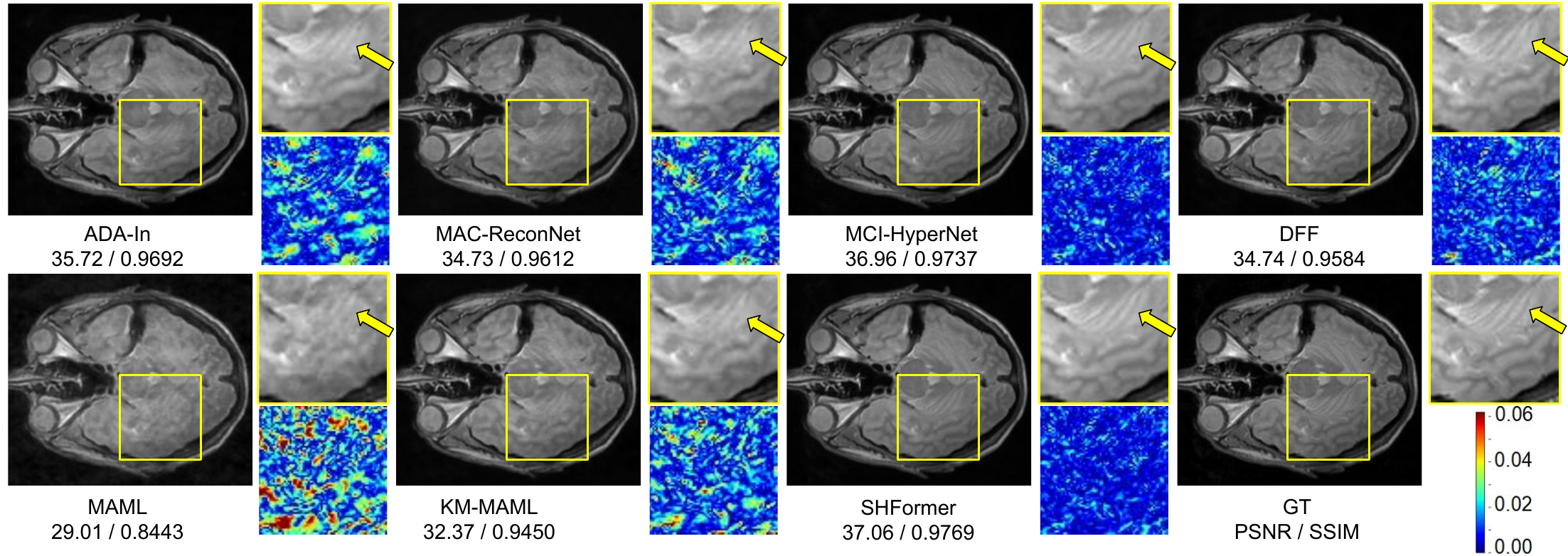}
    \caption{Qualitative comparison of the open-set generalization capabilities of SHFormer with other adaptive reconstruction models trained on T1, T2, and FLAIR and evaluated on PD. Here PD show much finer changes in the intensity level which are much better in the predictions of SHFormer, which is further affirmed with least residual errors.
    }
    \label{fig:opensetpd}
\end{figure*}

\begin{figure*}[t!]
    \centering
    \includegraphics[width=\linewidth]{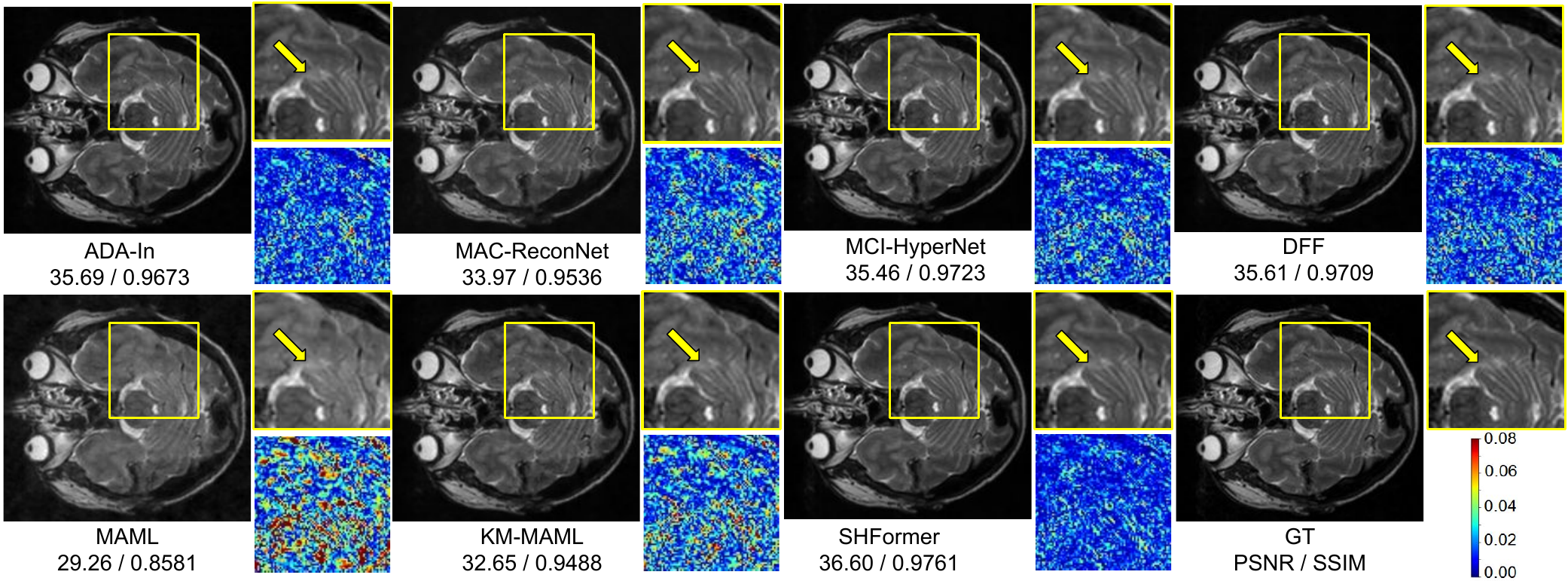}
    \caption{Qualitative comparison of the open-set generalization capabilities of SHFormer with other adaptive reconstruction models trained on T1, PD, and FLAIR and evaluated on T2. The residual images and the highlighted regions show improved recovery of faint structures for the proposed method over other methods.
    }
    \label{fig:opensett2}
\end{figure*}

\begin{figure*}[t!]
    \centering
    \includegraphics[width=\linewidth]{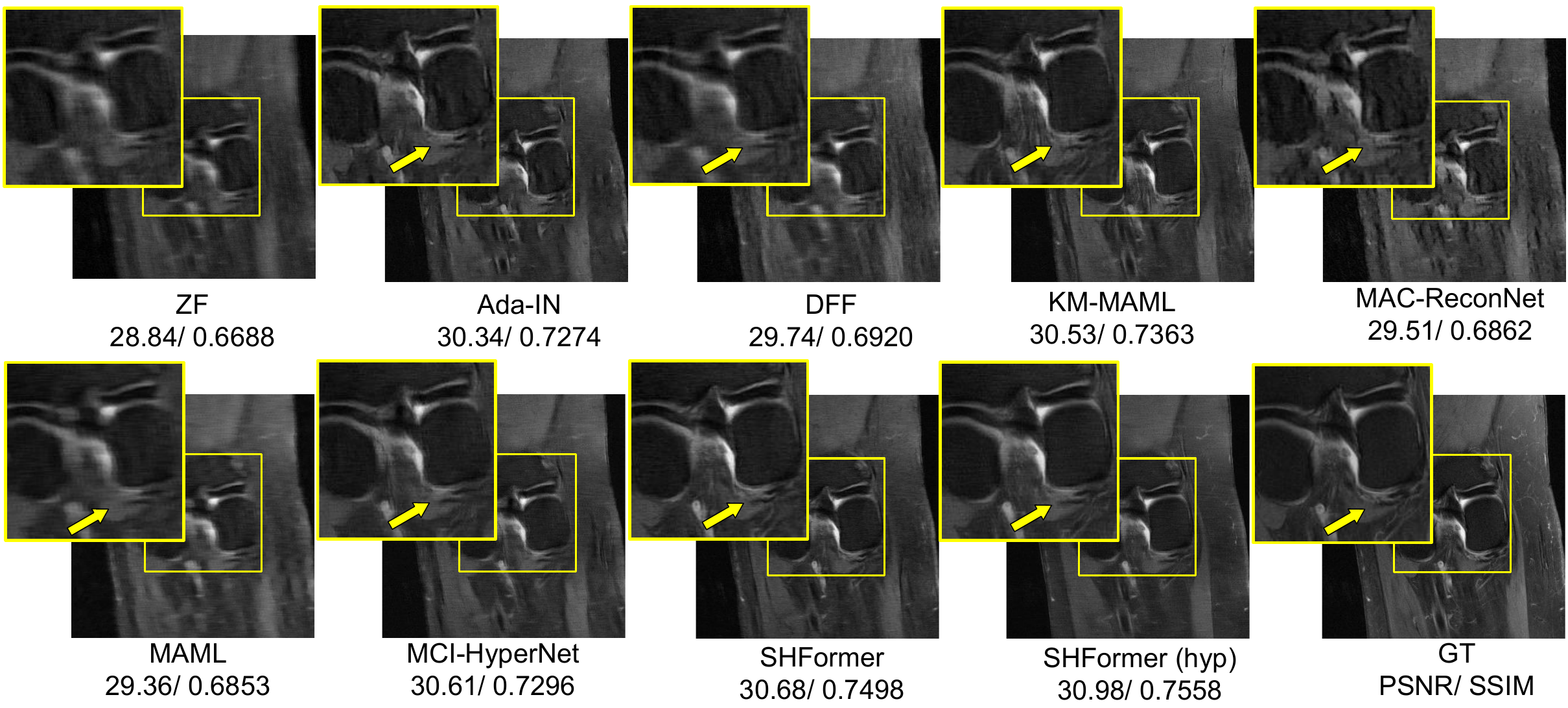}
    \caption{Qualitative comparison of the open-set generalization capabilities of SHFormer with other adaptive reconstruction models for complex-valued fastMRI dataset with coronal PD and PDFS modalities. As pointed out by yellow arrows, the highlighted regions indicate better reconstruction quality.
    }
    \label{fig:openset_fastmri}
\end{figure*}
\clearpage
\subsubsection{Open-set Generalization}

To understand the generalization capabilities of SHFormer further, we evaluate the models for an open-set generalization scenario of modality data drift using multi-contrast MRI data. We train  
the model on multiple MRI modality sequences and test on an unseen modality.

 
With this notion, we train combinations of T1, T2, FLAIR, and PD with various types and amounts of undersampling on a leave-one-out basis and evaluate the left-out modality as the target data domain, which deviates from the source domains. 

Table \ref{tab:openset} shows the quantitative comparison of SHFormer with other adaptive methods for three modality drift scenarios with longer scan time sequences, namely T2, FLAIR, and PD sequences. From the table, we see that the proposed model with DCT-based spectral attention is effective as compared to the competitive Fourier transform-based spectral filtering baseline model, DFF.
The proposed model outperforms task-level adaptive models - MAC-ReconNet and MCI-HyperNet and show significant improvement over Ada-IN. The improvements and competitiveness with DFF, the spectral baseline emphasizes the generalization ability of spectral filtering.  
The visual results in Figures  \ref{fig:opensetpd} and \ref{fig:opensett2} show that the proposed method recovers essential details and image textures much better compared to other approaches.  The residual errors for SHFormer with respect to the ground truth show minimum error in the region of interest as compared to other methods.
The proposed method selects mode-specific frequency components in the SF block of the CNN and adaptively rescales the features using the multi-level AM of the transformer with self-attention and dynamic high-pass filtering.
We also note that the proposed model learns spatially adaptive filters both in the frequency and image domains which combines both content and style, enhancing the robustness of the model. The frequency-space filtering dynamically modulates modality-specific frequency components during training to learn cross-domain generalizable features focusing on the style, while the high-pass filtering block learns to focus on the domain-invariant content.

\textbf{Analysis using complex-valued dataset: } 
Table \ref{tab:multimodal_complex_adapt} shows three cases - "Seen", "Unseen" and "Unseen (Adapt)" for the complex-valued fastMRI knee dataset. The "Seen" case involves training on coronal PD and PDFS and testing on the same two modalities. The two unseen scenarios showcase adaptability when the model trained on coronal PD knee is evaluated on two unseen cases.  (i) "Unseen" case - the model is directly evaluated on coronal PDFS knee without finetuning (ii) "Unseen (adapt)" case - the model is fine-tuned using one coronal PDFS knee patient volume in few gradient steps (30 steps). Two variants are considered for finetuning the proposed model - SHFormer where all the parameters are finetuned and SHFormer (hyp) where only the two hypernetworks used for the spectral and spatial AM, that form 2 - 3\% of the parameters are fine-tuned. In both "Unseen" and "Unseen (adapt)" cases, the model exhibits better adaptability with best reconstruction metrics as compared to other methods. The improvement margin in SSIM for the "Unseen (adapt)" case is $\sim$ 0.009 for SHFormer and $\sim$ 0.005 for SHFormer (hyp) over the "Unseen" case. We note that SHFormer (hyp) reuses the frozen pre-trained weights as much as possible and minimizes the pretrain-finetune architecture inconsistency \cite{glid}. The proposed hybrid AM showcases strong meta-learning capabilities while preserving the reconstruction quality (Figure \ref{fig:openset_fastmri}).

\begin{table}[]
\scriptsize
\centering
\caption{Quantitative comparison of SHFormer with other multimodal integrated architectures proposed for MRI}
\label{tab:multimodal_integrated}
\setlength{\tabcolsep}{4pt}
\renewcommand{\arraystretch}{1.5}
\begin{tabular}{ccccc}
\hline
\multirow{2}{*}{Method} & \multicolumn{2}{c}{4x}           & \multicolumn{2}{c}{5x}           \\ \cline{2-5} 
                        & PSNR           & SSIM            & PSNR           & SSIM            \\ \hline
MTrans                  & 33.70 $\pm$ 0.30 & 0.9201 $\pm$ 0.00 & 31.87 $\pm$ 0.68 & 0.8764 $\pm$ 0.01 \\
T2-Net                  & 33.93 $\pm$ 0.50 & 0.9196 $\pm$ 0.00 & 31.92 $\pm$ 0.50 & 0.8618 $\pm$ 0.01 \\
MI-Net                  & 34.96 $\pm$ 0.24 & 0.9391 $\pm$ 0.01 & 32.97 $\pm$ 0.65 & 0.8947 $\pm$ 0.02 \\
SHFormer                & 35.38 $\pm$ 0.28 & 0.9435 $\pm$ 0.02 & 34.02 $\pm$ 0.69 & 0.8955 $\pm$ 0.01 \\ \hline
\end{tabular}
\end{table}

\begin{figure}[t!]
    \centering
    \includegraphics[width=0.9\linewidth]{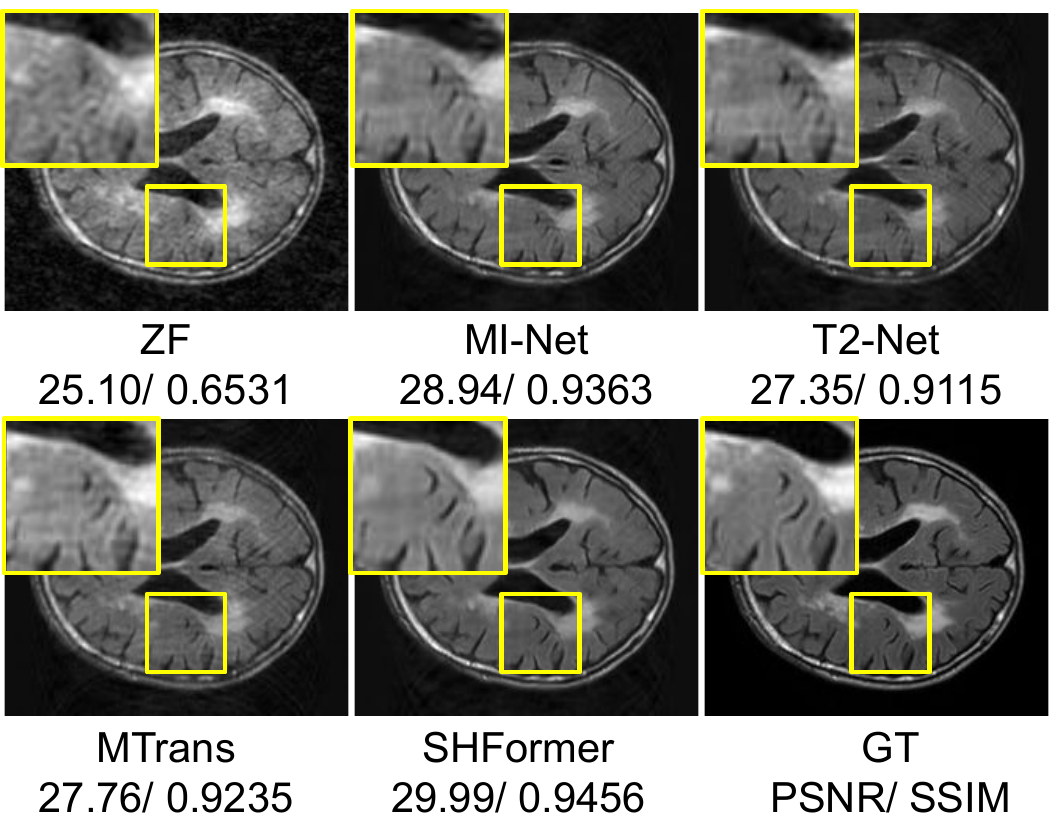}
    \caption{Qualitative comparison of SHFormer with other multimodal integrated methods for T2 FLAIR MRI reconstruction.
    }
    \label{fig:multimodal_integrated}
\end{figure}

\textbf{Comparative studies against other multimodal approaches:}
We have compared the proposed model against the three architectures - MTrans \cite{mtrans}, T2Net \cite{t2net} and MI-Net \cite{minet} for T2 FLAIR brain MRI reconstruction where MTrans and MINet consists of an auxiliary branch for T1 MRI reconstruction and assistance, respectively.
Table \ref{tab:multimodal_integrated} and Figure \ref{fig:multimodal_integrated} show the quantitative and qualitative comparison against these methods. From the table, we see that the proposed model outperforms other multimodal methods with a best improvement margin of $\sim$1 dB in PSNR for the 5x acceleration case. MI-Net is a competitive model concurring with our perspectives of cascaded attention modules and feature modulation. MI-Net consists of spatial domain attention modules while the proposed model provides hybrid attention in the spectral and spatial domains.

\begin{figure*}[t!]
    \centering
    \includegraphics[width=0.8\linewidth]{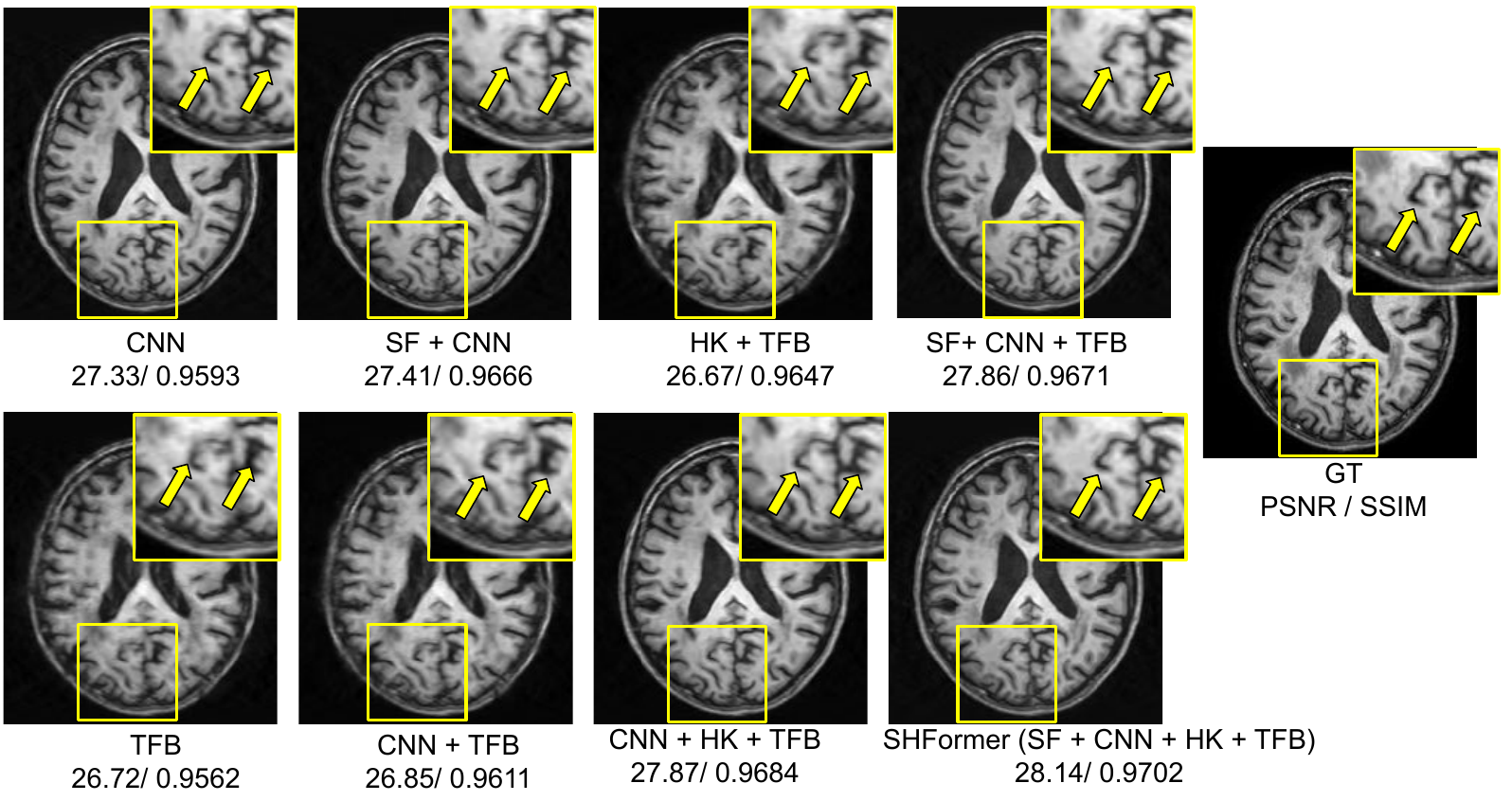}
    \caption{Ablative study of the SHFormer showing the contributions of various functional units of the architecture towards the overall reconstruction performance. The visual results indicated by the highlighted regions clearly show the importance of the design that combines the spectral filtering infused CNN block and the dynamic high-pass filtering transformer.
    }
    \label{fig:ablative}
\end{figure*}

\begin{table}[t!]
\scriptsize
\centering
\caption{Ablative Studies showing various combinations with and without the SF and HF modules to understand their contribution towards image reconstruction. Here "CNN" denotes the CNN layers of the method, "TFB" denotes the transformer block, "SF" denotes the spectral filtering module, and "HK" denotes the dynamic high-pass kernel generation block.}
\label{tab:ablative}
\setlength{\tabcolsep}{4pt}
\renewcommand{\arraystretch}{1.5}
\begin{tabular}{lcc}
\hline
\multicolumn{1}{c}{Method} & PSNR             & SSIM              \\ \hline
CNN                        & 40.21 $\pm$ 2.63 & 0.9674 $\pm$ 0.01 \\
TFB                        & 39.69 $\pm$ 2.00 & 0.9752 $\pm$ 0.00 \\
CNN + TFB                  & 39.48 $\pm$ 2.06 & 0.9760 $\pm$ 0.01  \\
HK + TFB                  & 39.86 $\pm$ 1.99 & 0.9778 $\pm$ 0.01 \\
SF + CNN                   & 40.54 $\pm$ 2.90 & 0.9759 $\pm$ 0.01 \\
SF + CNN + TFB             & 41.19 $\pm$ 2.97 & 0.9824 $\pm$ 0.02 \\
CNN + HK + TFB            & 40.97 $\pm$ 2.88 & 0.9793 $\pm$ 0.01 \\
SF + CNN + HK + TFB       & 41.32 $\pm$ 3.08 & 0.9841 $\pm$ 0.01 \\ \hline
\end{tabular}
\end{table}

\begin{figure*}[t!]
    \centering
    \includegraphics[width=0.9\linewidth]{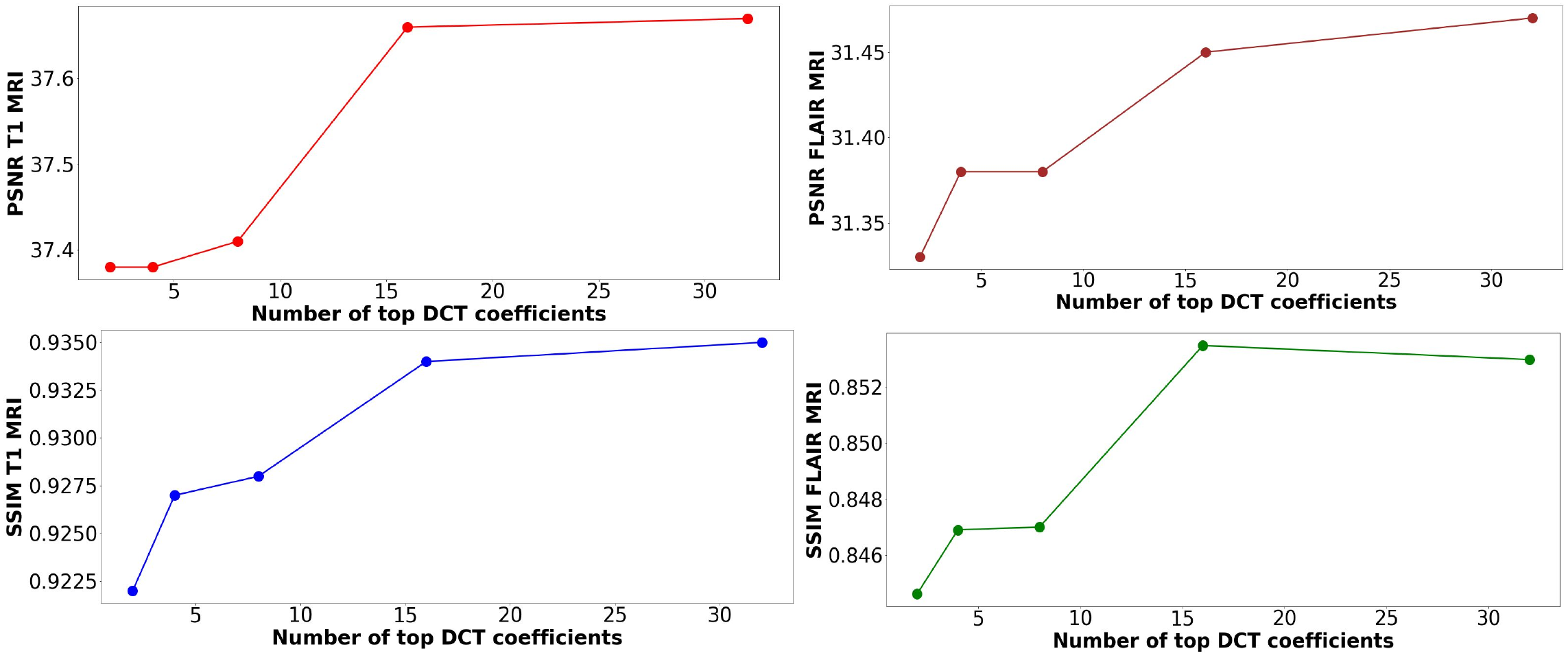}
    \caption{Ablative study on varying numbers of significant DCT components in the spectral attention module. The performance increases as the number of frequency components increase and saturates beyond a certain extent, indicating redundancy in the parameters.
    }
    \label{fig:dct_num_coeff}
\end{figure*}

\subsubsection{Ablative Studies}
\textbf{Ablation study on the different functional modules: } To understand the contribution of the proposed DCT-based SFCNN and the HKTFB sub-networks towards the reconstruction performance, we perform an ablative study with eight different functional combinations of the various modules of the proposed approach, as shown in Table \ref{tab:ablative}.

In the table, CNN denotes no AM, wherein the network consists of only the vanilla CNN layers without the transformer, spectral filtering (SF), and dynamic high-pass kernel generation (HK) blocks. The row, "TFB" denotes the presence of only the vanilla transformer network without the HK layer. The row, "CNN + TFB" denotes the combination of vanilla CNN and transformer blocks. The row  "HK + TFB" denotes the presence of the transformer network with the dynamic HK block without the CNN.  
The row "SF + CNN" denotes the case with only the CNN with SF module without the transformer network. "SF + CNN + TFB" denotes the presence of SFCNN and the transformer without the dynamic HK layer. 
"CNN + HK + TFB" denotes the presence of the CNN (without SF) with the HPF transformer. The last row indicates the presence of all the functional modules of the proposed approach. From the table, our observations are: (i) The proposed approach combining CNN with SF and HK transformer is essential for better accuracy (ii) The cases with only CNN or TFB or both perform relatively less. (iii) The spectral filtering operation ("SF + CNN" case) within the CNN block enhances the accuracy by margins of 0.3 dB in PSNR and 0.01 in SSIM suggesting the importance of global spectral attention weight prediction (iii) The high-pass filtering driven transformer ("HK + TFB" case) clearly shows improvements in SSIM by margins of over 0.01 as compared to the CNN baseline and better metrics than the "SF + CNN" case. This indicates that the proposed HF transformer improves the high-frequency learning capabilities of the transformer significantly (iv) The cases  "SF + CNN + TFB" and "CNN + HK + TFB" indicate the importance of combining the inductive bias of either the SFCNN or the HKTFB, providing local and global contextual learning, respectively. (v) The proposed architecture (SF + CNN + HK + TFB) combines the benefits of robustness and contextual learning. This observation concurs with the qualitative results in Figure \ref{fig:ablative} showing better reconstruction quality over other modular combinations.

\textbf{Ablation study on the number of spectral components:}
We further perform ablation studies based on varying numbers of top DCT components of the spectral attention module as shown in Figure \ref{fig:dct_num_coeff} for two modalities - T1 and FLAIR MRI. The figure shows the PSNR and SSIM metrics for varying numbers of DCT coefficients - 2, 4, 8, 16, and 32. We note that as the number of spectral components increases, the performance boosts up. However, increasing the number of coefficients further increases the number of weights of the spectral AM layer without significant performance improvements. A surplus number of parameters might not learn reusable features to further improve the reconstruction fidelity. In all our experiments, we used 16 frequency components.



\begin{figure}[t!]
    \centering
    \includegraphics[width=\linewidth]{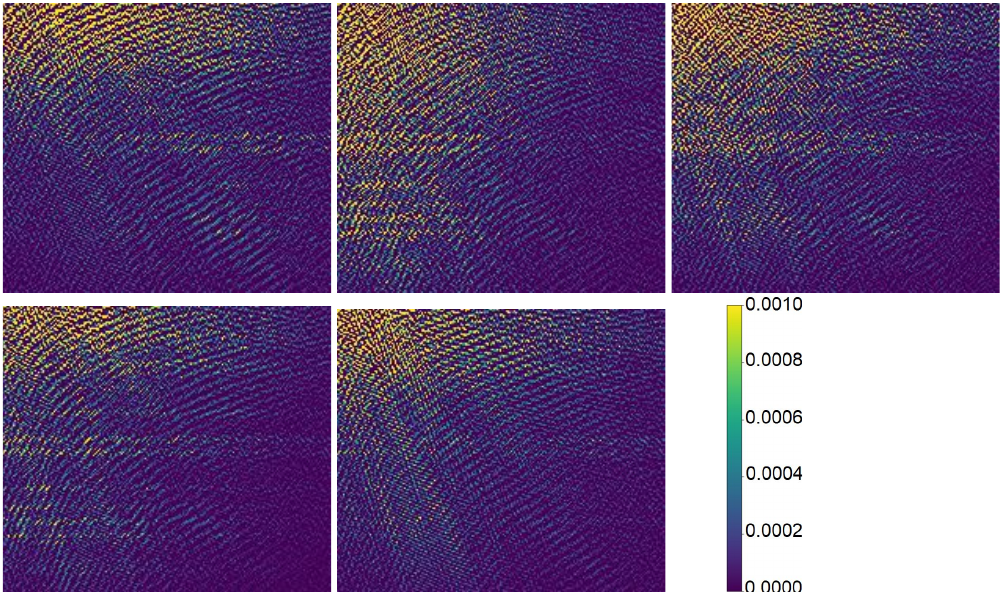}
    \caption{Visualization of the learned DCT-based attention maps for spectral attention. The DCT coefficients at the top left position are dense and correspond to low and mid frequency components while the maps at the lower right are sparse and indicate high frequency components. 
    }
    \label{fig:dctviz}
\end{figure}

\begin{figure*}[t!]
    \centering
    \includegraphics[width=\linewidth]{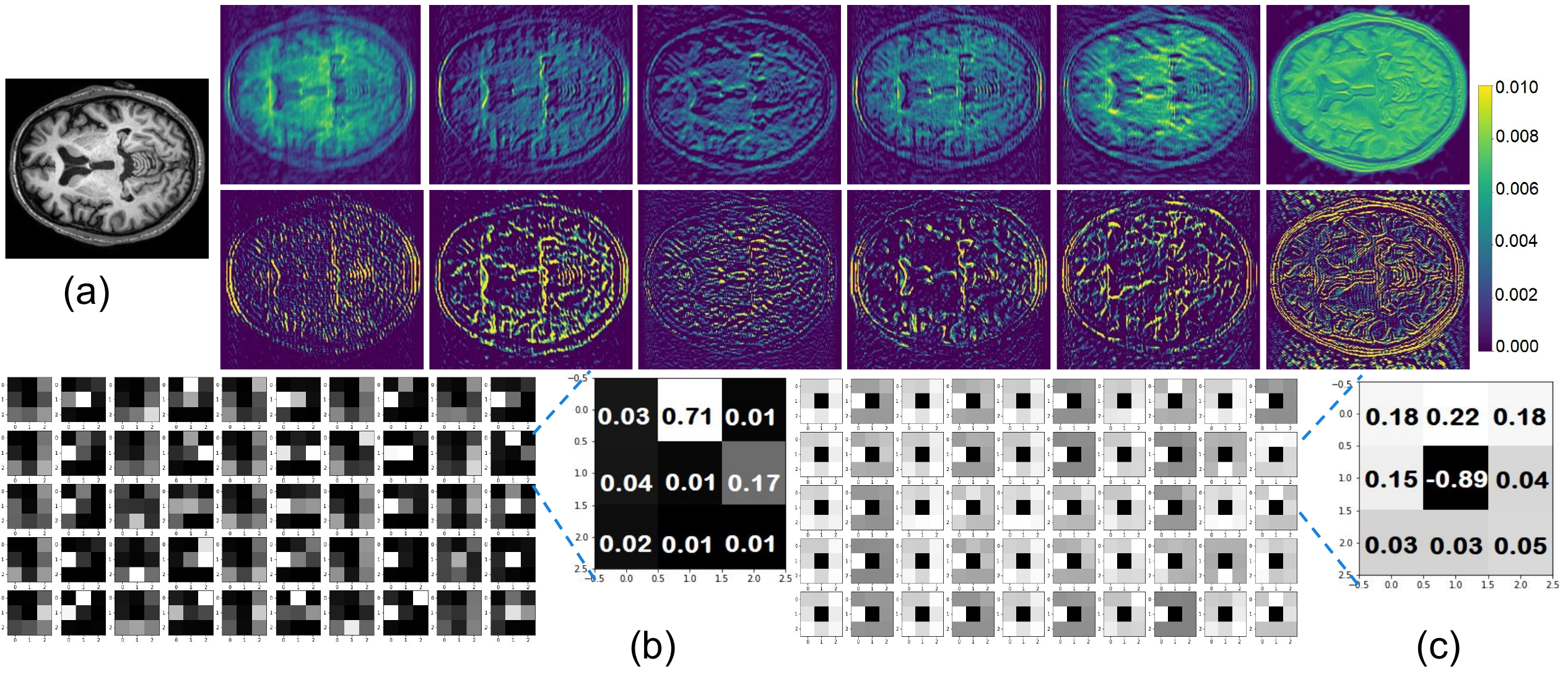}
    \caption{(a) Visualization of the low pass filtered features (top) and features of the dynamic high pass filtering block (bottom). The features show higher response in regions where the intensity transitions are sharp, indicating that the dynamic kernels learns to filter high-frequency details of the feature maps. (b) Visualization of the learned low pass kernels, each kernel summing to one (c) Visualization of the learned high pass kernels, each summing to zero.
    }
    \label{fig:hpfviz}
\end{figure*}

\subsubsection{Interpretation of the Learned Features}

To understand the contributions of the proposed DCT-based spectral filtering module in the CNN and the dynamic high-pass filtering block in the transformer,  we visualize the learned spatial masks of the SF module and the self-attentive features of the transformer after convolving the HF filters with the features. 
Figure \ref{fig:dctviz} shows the DCT-based spectral attention masks. Our observations from the figure are, (i) the spectral domain feature maps show DCT coefficients with higher responses in the top left corner, indicating the SF block enhances relatively low-frequency components while keeping the high-frequency components small. This observation is in accordance with the F-principle \cite{TrainbehaviourDNNFreq} which states that deep neural networks first capture the less noisy low-frequency components of the training data and this helps to learn transferable features to generalize to deviations in the data during inference. (ii) The feature maps exhibit variations in responses with some showing higher response along vertical components and some along the horizontal frequency components, indicating that the spectral filtering can pick a range of frequency modulations in the latent space.


Figure \ref{fig:hpfviz} (a) shows the output low-pass (top) and high-pass (bottom) filtered feature representations for the axial T1 brain input image showing the ability to capture and focus on the high frequencies. The low-pass filtered features exhibit higher response for homogeneous regions.
The high-pass filtered features exhibit higher response for high-frequency regions namely edges, sharp intensity transitions at various orientations and lower response for homogeneous regions, where the image gradients are small. Figure \ref{fig:hpfviz} (b) and (c) show the visualizations of the low-pass and high-pass kernels respectively.


\FloatBarrier
\clearpage
\section{Conclusion}
In this paper, we have introduced frequency perspectives in attention mechanisms to achieve adaptability in MRI reconstruction across heterogeneous data scenarios, namely, multiple contrasts and under-sampling mask patterns and levels. Our experiments validate the robustness of the proposed approach to a wide range of acquisition settings, showcasing closed-set and open-set generalization capabilities against previous (i) adaptive deep learning methods based on dynamic weight prediction, and meta-learning at the task level, and (ii) instance normalization, spatial and spectral attention mechanisms at the instance level. We further showcase the validity and efficiency across learning modes - supervised and self-supervised, and generative model setting. The qualitative results show that our model exhibits superior recovery of high-frequency information, enabling enhanced reconstruction of structures. We are currently extending these techniques to other types of MRI images, like dynamic contrast-enhanced images for other MRI imaging tasks.

\bibliographystyle{cas-model2-names}

\bibliography{cas-refs}





\end{document}